\documentclass[11pt,a4paper]{article}
\usepackage[left=2.5cm,right=2.5cm,top=2.5cm,bottom=2.5cm]{geometry}
\usepackage{color}

\usepackage{graphicx}
\usepackage{amsmath}
\usepackage{amsfonts}
\usepackage{amssymb}
\usepackage{bm}
\usepackage{pdflscape}
\usepackage{url}
\usepackage{bbold}
\usepackage{xcolor}
\usepackage{enumerate}
\usepackage{comment}
\usepackage{float}  
\usepackage[colorlinks=true, pdfstartview=FitV, linkcolor=blue, 
            citecolor=blue, urlcolor=blue]{hyperref}
\usepackage[version=3]{mhchem}

\newtheorem{remark}{Remark}

\newtheorem{definition}{Definition}[section]

\newcommand{\bitem}{\begin{itemize}}
\newcommand{\eitem}{\end{itemize}}
\newcommand{\mc}[1]{\mathcal{#1}}

\newcommand{\N}{\mathbb{N}}
\newcommand{\R}{\mathbb{R}}

\newcommand{\bpm}{\begin{pmatrix}}
\newcommand{\epm}{\end{pmatrix}}
\newcommand{\bvm}{\begin{vmatrix}}
\newcommand{\evm}{\end{vmatrix}}
\newcommand{\bsm}{\left(\begin{smallmatrix}}
\newcommand{\esm}{\end{smallmatrix}\right)}
\newcommand{\T}{\top}

\newcommand{\ol}[1]{\overline{#1}}
\newcommand{\wh}[1]{\widehat{#1}}
\newcommand{\wt}[1]{\widetilde{#1}}
\newcommand{\la}{\langle}
\newcommand{\ra}{\rangle}

\newcommand{\veps}{\varepsilon}

\newcommand{\w}{\omega}

\newcommand{\vphi}{\varphi}

\newcommand{\eins}{\mathbb{1}}

\DeclareMathSymbol{\mydiv}{\mathbin}{symbols}{"04}

\DeclareMathOperator{\tr}{tr}
\DeclareMathOperator{\Diag}{Diag}

\DeclareMathOperator{\vvec}{vec}

\DeclareMathOperator{\Span}{span}

\makeatletter
\def\widebreve{\mathpalette\wide@breve}
\def\wide@breve#1#2{\sbox\z@{$#1#2$}%
     \mathop{\vbox{\m@th\ialign{##\crcr
\kern0.08em\brevefill#1{0.8\wd\z@}\crcr\noalign{\nointerlineskip}%
                    $\hss#1#2\hss$\crcr}}}\limits}
\def\brevefill#1#2{$\m@th\sbox\tw@{$#1($}%
  \hss\resizebox{#2}{\wd\tw@}{\rotatebox[origin=c]{90}{\upshape(}}\hss$}
\makeatletter
\numberwithin{equation}{section}
\numberwithin{figure}{section}

\title{Learning System Parameters from Turing Patterns}
\usepackage{authblk}
\author[1]{David Schn{\"o}rr}
\author[2]{Christoph Schn{\"o}rr}
\affil[1]{Imperial College, London, UK}
\affil[2]{Heidelberg University, Germany}
\date{}

\begin{document}

\maketitle

\begin{abstract}
The Turing mechanism describes the emergence of spatial patterns due to spontaneous symmetry breaking in reaction-diffusion processes and underlies many developmental processes. Identifying Turing mechanisms in biological systems defines a challenging problem. 
This paper introduces an approach to the prediction of Turing parameter values from observed Turing patterns. The parameter values correspond to a parametrized system of reaction-diffusion equations that generate Turing patterns as steady state. The Gierer-Meinhardt model with four parameters is chosen as a case study. A novel invariant pattern representation based on resistance distance histograms is employed, along with Wasserstein kernels, in order to cope with the highly variable arrangement of local pattern structure that depends on the initial conditions which are assumed to be unknown. This enables to compute physically plausible distances between patterns, to compute clusters of patterns and, above all, model parameter prediction: for small training sets, classical state-of-the-art methods including operator-valued kernels outperform neural networks that are applied to raw pattern data, whereas for large training sets the latter are more accurate. Excellent predictions are obtained for  single parameter values and reasonably accurate results for jointly predicting all parameter values.
\end{abstract}

\textbf{keywords}: vector-valued parameter prediction, Turing patterns, resistance distance histograms 

\tableofcontents

\section{Introduction}\label{intro}

\subsection{Motivation and Overview}

Reaction-diffusion models in the form of Eq.~\eqref{eq:pde} are used to describe the dynamic behaviour of interacting and diffusing particles in various disciplines including biochemical processes \cite{murray2001mathematical}, ecology \cite{holmes1994partial}, epidemiology \cite{martcheva2015introduction} and tumor growth \cite{gatenby1996reaction}. Here, we are interested in systems where the interaction of particles can give rise to spontaneous symmetry breaking of a homogenous system by means of the so-called \textit{Turing mechanism} which was first described by Alan Turing in 1952 \cite{turing1952chemical}. It describes the scenario where a stable steady state of a non-spatial system of ordinary differential equations becomes unstable due to diffusion \cite{murray1982parameter,Pertham:2015aa}. This phenomeon is hence also referred to as \textit{diffusion-driven instability}. Such instabilities typically give rise to stable non-homogenous spatial patterns. In two spatial dimensions, for example, these patterns can take various forms such as spots, stripes or labyrinths \cite{murray2001mathematical}. This variety of patterns is generated by a  reaction-diffusion model that is parametrised by a few parameters that represent physical quantities of the system, such as reaction and diffusion rate constants. Certain versions of the well-known Gierer-Meinhardt model, for example, comprises four effective parameters (cf.~Section \ref{sec:examples}) \cite{gierer1972theory,murray2001mathematical}.  

It was not until almost four decades after Alan Turing's seminal work that the first experimental observation of a Turing pattern was realised in a chemically engineered system \cite{castets1990experimental}. Recently, as a first practical application, a chemical Turing system was engineered to manufacture a porous filter that can be used in water purification \cite{tan2018polyamide}. In biological systems,
Turing patterns are regarded as the main driving mechanism in the formation of spatial structures in various biological systems, including patterning of palatel  ridges  and  digits, hair follice distribution, feather formation, and patterns on the skins of animals such as fish and zebras \cite{economou2012periodic, raspopovic2014digit, sick2006wnt, jung1998local, nakamasu2009interactions}. For a recent survey, see \cite{landge2020pattern}. However, the biological and mathematical complexity has often prevented identification of the precise molecular mechanisms and parameter values underlying biological systems. One difficulty in fitting models to experimental data stems from the high sensitivity of the arrangement of local structure on the initial conditions that are usually unknown in practice. 

In this paper, we focus on the \textit{inverse} problem: given a non-homogenous spatial pattern and a reaction-diffusion model, the task is to predict the parameter values that generate the pattern as steady state of the reaction-diffusion equation. To this end, we introduce a novel pattern representation in terms of resistance distance histograms that effectively represents the spatial structure of patterns, irrespective of the local variability stemming from different initial conditions. This enables to compute almost invariant distances between patterns, to compute clusters of patterns and, above all, to predict model parameter values.

Specifically, we focus on the Gierer-Meinhardt model as a case study and apply and compare state-of-the-art machine learning techniques and  neural network architectures for prediction. The former approaches comprise nonlocal image features for the invariant representation of spatial patterns and kernel-based methods for parameter prediction. The latter neural networks are either applied using the afore-mentioned nonlocal image features, or are directly applied to the raw pattern data and perform feature extraction and parameter prediction simultaneously. All predictors are trained using various parameter values that generate 
diffusion-driven instabilities 
and corresponding spatial patterns that result from solving numerically the reaction-diffusion equations. 

\subsection{Related Work}\label{sec:Related-Work}
The problem studied in this paper, parameter prediction from observed Turing patterns, has been studied in the literature from various angles. We distinguish three categories and briefly discuss few relevant papers.
\begin{description}
\item[Turing parameter estimation by optimal control.] 
The work \cite{garvie2010efficient} presents an approach for estimating parameter values by fitting the solution of the reaction-diffusion equation to a given spatial pattern. This gives rise to a PDE-constraint problem of optimal control requiring sophisticated numerics; see also \cite{stoll2016fast}. A similar approach is studied in \cite{sgura2019parameter}. 

The authors of \cite{garvie2010efficient} show and demonstrate that the proposed control problem is solvable which indicates that the task studied in our work, i.e.~learning \textit{directly} the pattern-to-parameter mapping, is not unrealistic. The approach of \cite{garvie2010efficient} has been generalized by \cite{garvie2014identification} in order to handle also non-constant spatially-distributed parameters. In our work, we only consider constant parameter values.

A strong property of approaches employing PDE-based control is that they effectively cope with \textit{noisy} observed patterns, provided that the type of noise is known such that a suitable objective function can be set up. A weak point is that in some of these papers the initial conditions are assumed to be known which is not the case in realistic applications, and that sophisticated and  expensive numerics is required.

The problem to control PDEs that generate \textit{time-varying} travelling wave patterns has been studied recently \cite{uzunca2017optimal,karasozen2020reduced,shangerganesh2020optimal}. In these works the focus is on fitting the trajectory of the evolving pattern in function space, however, rather than on estimating model parameter values that are assumed to be given.

\item[Turing parameter estimation by statistical inference.] 
The paper \cite{campillo2019bayesian} presents a Bayesian approach to parameter estimation using the reaction diffusion equation as forward mapping and a data likelihood function corresponding to additive Gaussian zero mean noise superimposed on observed patterns. Given a pattern, the posterior distribution on the parameter space is explored using expensive MCMC computations. A weak point of this approach shared with the works discussed above in the former category is that the initial conditions are assumed to be known. This assumption is not required in our approach presented below.

Closer to our work is the recent paper \cite{kazarnikov2020statistical}. These authors also study model parameter identification from steady-state pattern data only, without access to initial conditions or the transient pattern evolving towards the steady state. The key idea is to model statistically steady-state patterns of `the same class', i.e.~patterns whose spatial structure differ considerably due to different initial conditions. This is achieved by adopting a Gaussian model for the empirical distribution of discretized $L_{2}$ distances between spatial patterns, which can be justified theoretically in the large sample limit. This approach requires a few dozen to hundreds of novel test patterns to estimate model parameters.

In our work, we proceed differently: an almost invariant representation of patterns of `the same class' is developed. This is advantageous in practice since parameter prediction can be done for each \textit{single} novel test pattern.

\item[Turing parameter estimation: other approaches.] 
The work \cite{murphy2018parameter} focus on the identification of parameter values through a linear stability analysis on various irregular domains, assuming that the corresponding predicted pattern is close to a desired or observed pattern. However, the authors admit that, in many cases, the steady-state pattern may \textit{not} be an eigenfunction of the Laplacian on the given domain, since the nonlinear terms play a role in the resultant steady-state pattern. In our work, we solely focus on steady-state patterns and the inverse parameter value prediction problem.

A recent account of the broad variety of Turing pattern generating mechanisms and corresponding identifiability issues is given by \cite{woolley2021bespoke}. In our work, we focus on the well-known Gierer-Meinhardt model and study the feasibility of predicting points in the four-dimensional parameter space based on given steady-state patterns.
\end{description}

\subsection{Contribution and Organisation}
We introduce a novel representation of the spatial structure of Turing patterns which is achieved by computing \textit{resistance} distances \textit{within each} pattern, followed by discretization and using the empirical distribution of resistance distances as class representative. Discretization effects are accounted for by using the Wasserstein distance and a corresponding kernel function for comparing \textit{pairs} of patterns. Based on this representation we 
present results of a feasibility regarding the prediction of model parameter values from observed patterns. To our knowledge, this is the first paper that applies machine learning methods to the problem of mapping directly Turing patterns to model parameter values of a corresponding system of reaction-diffusion equations that generate the pattern as steady state. Adopting the Gierer-Meinhardt model as a case study, we demonstrate that about 1000 data points suffice for highly accurate prediction of single model parameter values using state-of-the-art kernel-based methods. The accuracy decreases for predictions of all four model parameter values but is still sufficiently good in terms of the normalized root-mean-square error and the corresponding pattern variation. In the large data regime ($\geq 20.000$ data points) predictions by neural networks trained directly on raw pattern data outperform kernel-based methods.

Our paper is organized as follows. Section \ref{sec:Turing-definition} summarizes basics of Turing patterns that are required in the remainder of the paper: definition of diffusion-driven patterns; discretization and a numerical algorithm for solving a system of semi-linear reaction diffusion equations whose steady states correspond to the patterns that are used as input data for model parameter prediction; the Gierer-Meinhardt PDE and its parametrization. Section \ref{sec:Feature-extraction} details the features that are extracted from spatial patterns in order to predict model parameter values. A key feature are histograms of resistance distances that represent spatial pattern structure in a proper invariant way. Section \ref{sec:Learning-Patterns} introduces four methods for model parameter prediction from observed patterns: two kernel-based methods (basic SVM regression and operator-valued kernels) and neural networks are applied to either nonlocal pattern features or to the raw pattern data directly. Numerical results are reported and discussed in Section \ref{sec:Experiments}. We conclude in Section \ref{sec:Conclusion}.

\subsection{Basic Notation}
We set $[n]=\{1,2,\dotsc,n\}$ and $\eins_{n}=(1,\dotsc,1)^{\T}\in\R^{n}$ for $n\in\mathbb{N}$. The Euclidean inner product is denoted by $\langle p,q\rangle$ for vectors $q, p \in \mathbb{R}^{n}$ with corresponding norm $\|q\|=\sqrt{\langle q,q\rangle}$. The $\ell^{\infty}_{n}$-norm is denoted by $\|p\|_{\infty} = \max\{|p_{i}|\colon i\in[n]\}$.  $\la A, B\ra = \tr{(A^{\T}B)}$ with trace $\tr{(\cdot)}$ denotes the inner product for two matrices $A, B$.
The spectral matrix norm is written as $\|A\|$. The symbol $\lambda$ with matrix argument denotes an eigenvalue $\lambda(A)$ of the matrix. $\mathbb{R}^{n}_{+}$ is the nonnegative orthant and $u>0$ means $u_{1}>0,\dotsc,u_{n}>0$ if $u \in \R^{n}$. $\Diag(u)$ is the diagonal matrix that has the components of a vector $u$ as entries. Similarly, $\Diag(A_{1},\dotsc,A_{n})$ is the block diagonal matrix with matrices $A_{i},\,i\in [n]$ as entries. The probability simplex is denoted by $\Delta_{n}=\{p \in \R_{+}^{n}\colon \la\eins_{n},p\ra=1\}$.

\section{Turing Patterns: Definition and Computation}\label{sec:Turing-definition}
This section provides the required background on Turing patterns: reaction-diffusion systems (Section \ref{sec:Reaction-diffusion}), Turing instability and patterns (Section \ref{sec:Turing-patterns}), and a numerical algorithm for computing Turing patterns (Section \ref{sec:numerical_simulation}). We refer to,  e.g., \cite{murray2001mathematical, Pertham:2015aa} for comprehensive expositions and to \cite {kondo2010reaction, landge2020pattern} for recent reviews. \

\subsection{Reaction-Diffusion Models}\label{sec:Reaction-diffusion}

Consider a system of $N$ interacting species described by the state vector $u(t)=(u_1(t),\ldots, u_N(t))$, where $u_i(t) \in \mathbb{R}_+$ is the time-dependent concentration of the $i$th species. We assume that the dynamics is governed by an autonomous system of ordinary differential equations
\begin{equation}\label{eq:ode}
  \frac{d}{dt} u(t) 
  = 
    f (u(t)), \quad u(0) = u_{0} > 0,
\end{equation}
where initial condition $u_{0} \in \mathbb{R}_{+}^{N}$ is assumed to be positive and $f\colon\mathbb{R}^{N}\to\mathbb{R}^{N}$ encodes interactions of the species. We further assume that the functions $f_{i},\,i \in [N]$ are continuously differentiable with bounded derivatives. `Autonomous' means that $f$ does not explicitly depend on the time $t$. 

Next, model \eqref{eq:ode} is extended to a spatial scenario including diffusion. Concentrations $u_{i}(t),\,i\in [N]$ are replaced by space-dependent concentration fields $u(r, t)=(u_1(r, t), \dotsc, u_N(r, t))$, where $r=(r_1, \dotsc, r_M)$ denotes a point in a region $S \subset \mathbb{R}^M$. The dynamics of these fields is described by a system of reaction-diffusion equations
\begin{subequations}\label{eq:pde}
\begin{align}
  \frac{\partial}{\partial t} u(r, t) 
  & = 
    f(u(r,t)) + D\Delta_{N} u(r, t), \\
    u(r,0) &= u_{0}(r),\quad r \in S, \quad t \geq 0,
    \intertext{where}
    D&= \Diag(\delta_{1},\dotsc,\delta_{N}) \in \R^{N\times N} \\
    \Delta_{N} &= \Diag(\Delta,\dotsc,\Delta)
\end{align}
\end{subequations}
with diffusion constants $\delta_i \in \mathbb{R}_+,\,i\in [N]$ of species $i \in [N]$. $\Delta_{N}$ denotes the block-diagonal differential operator that separately applies the ordinary Laplacian $\Delta = \partial^2/\partial r_1^2 + \dotsb +  \partial^2/\partial r_M^2$ to each component function $r\mapsto u_{i}(r,t),\,i\in [N]$. 

System \eqref{eq:pde} has to be supplied with boundary conditions in order to be well-posed. A common choice are homogeneous Neumann conditions. We choose \textit{periodic} boundary conditions, however, because this considerably speeds up the generation of training data by numerical simulation (Section \ref{sec:numerical_simulation}), yet does \textit{not} facilitate or change in any essential way the \textit{learning problem} studied in this paper.

\subsection{Turing Patterns}\label{sec:Turing-patterns}
We characterize \textit{Turing instabilities} that cause \textit{Turing patterns}.
Suppose  $u^* \in \mathbb{R}_{+}^{N}$ is an equilibrium point of \eqref{eq:ode} satisfying $f(u^{\ast})=0$. In order to assess the stability of  $u^*$, we write
\begin{equation}
  u(t) 
  =
    u^* + \epsilon\,\wt{u}(t), \quad \epsilon > 0, \quad \wt{u}(t) \in \mathbb{R}^N
\end{equation}
and compute a first-order expansion of the system \eqref{eq:ode},
\begin{equation}\label{eq:ode_expanded}
  \frac{d}{dt}\wt{u}(t) 
  =
    J(u^{\ast}) \wt{u}(t) + O(\epsilon),
\end{equation}
with the Jacobian 
\begin{equation}\label{eq:non-spatial_jacobian}
  J(u^{\ast}) = \big(J_{i,j}(u^{\ast})\big)_{i,j\in [N]},\qquad
  J_{i,j}(u^{\ast})
  =
    \frac{\partial f_i(u)}{\partial u_j}\bigg|_{u=u^*}. 
\end{equation}
Let $\lambda_1=\lambda_{1}(J(u^{\ast})), \dotsc, \lambda_N=\lambda_{N}(J(u^{\ast})) \in \mathbb{C}$ be the eigenvalues of $J$ at $u^{\ast}$. The equilibrium $u^*$ is \textit{asymptotically stable} if and only if $\text{Re}(\lambda_i)<0, i \in [N]$ \cite[Cor.~6.1.2]{Schaeffer:2016aa}, that is a region of attraction $U(u^{\ast})$ containing $u^{\ast}$ exists such that $u(t)\in U(u^{\ast})$ implies $u(t) \to u^{\ast}$ as $t \to \infty$.

Assuming that $u^*$ is asymptotically stable, we next consider the extended system  \eqref{eq:pde} that involves spatial diffusion. Let $u^{\ast} = u^{\ast}(r)$ denote the spatially constant extension of the equilibrium point that neither depends on the time $t$ nor on the spatial variable $r$: $u^{\ast}(r,t)=u^{\ast}(r)=u^{\ast}(r'),\;\forall r, r' \in S$. Hence $\Delta_{N} u^* =0$. Due to the diffusion terms, this equilibrium of $f$ may not be stable anymore for the system \eqref{eq:pde}, however. 
To assess the stability of $u^{\ast}$, a linear stability analysis is conducted using the ansatz
\begin{equation}\label{eq:spatial_perturbation}
  u(r,t) 
  = 
    u^* + \epsilon\,\wt{u}(t) e^{i \la q, r\ra},
\end{equation}
where $i=\sqrt{-1}$, $\epsilon \in \mathbb{R}_+$, $\wt{u}(t) \in \mathbb{R}^N$, $q \in \mathbb{R}^M$. The perturbation $\wt{u}(t) e^{i \la q, r\ra}$ conforms to the eigenvalue problem of the \textit{linearized} spatial system \eqref{eq:pde}. 
Substituting this ansatz into \eqref{eq:pde} and expanding to the first order with respect to $\epsilon$ yields a linear system of equations for $\wt{u}(t)$, similar to Eq.~\eqref{eq:ode_expanded}, but with Jacobian $\wt{J}$ given by 
\begin{equation}\label{eq:spatial_jabobian}
  \wt{J}(u^{\ast},q)
  =
    J(u^{\ast}) - |q|^2 D.
\end{equation}
 Let $\wt{\lambda}_1(q)=\lambda(\wt{J}(u^{\ast},q)), \dotsc, \wt{\lambda}_N(q)=\lambda(\wt{J}(u^{\ast},q))$ be the eigenvalues of $\wt{J}(u^{\ast},q)$. For $\|q\|=0$, we have 
 $\wt{\lambda}_i(0) = \lambda_i,\,i \in [N]$ (eigenvalues of $J(u^{\ast})$ given by \eqref{eq:non-spatial_jacobian}) and hence $\text{Re}(\wt{\lambda}_i(0))<0, \,i\in [N]$, since $u^*=u^{\ast}(r)$ is equal to the equilibrium of non-spatial system \eqref{eq:ode} for every $r$.

\begin{figure}[t]
\centerline{
\includegraphics[width= 0.45\textwidth]{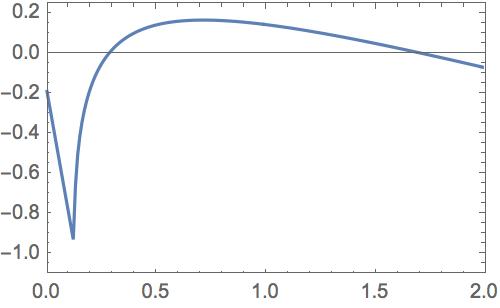}
\hfill
\includegraphics[width= 0.45\textwidth]{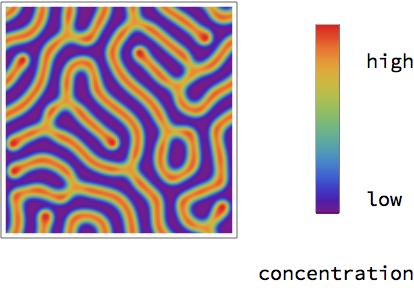}
}
\centerline{
\parbox{0.45\textwidth}{\centering (a)}\hfill
\parbox{0.45\textwidth}{\centering (b)}
}
\caption{(a) Panel (a) displays the eigenvalue $\wt{\lambda}_i(q)$ of the Jacobian $\wt{J}(u^{\ast},q)$ in Eq.~\eqref{eq:spatial_jabobian} that gives rise to the Turing instability. The real part of $\wt{\lambda}_i(q)$ is shown, evaluated at an asymptotically stable equilibrium point $u^{\ast}$ for the Gierer-Meinhardt model of Eq.~\eqref{eq:gm_model},
as a function of $\|q\|^2$. 
The parameters of the model were set to $a=0.01$, $b=1.2$, $c=0.7$, $\delta=40$ and the scaling parameter to $s=1$. One has $\text{Re}(\wt{\lambda}_i(q))|_{q=0}<0$ due to an asymptotically stable equilibrium as explained after Eq.~\eqref{eq:non-spatial_jacobian}. We find that $\text{Re}(\wt{\lambda}_i(q))$ becomes positive for an intermediate range of $\|q\|$ values, which indicates a Turing instability.
(b) The first species field $u_{1}(r,t)$ of the solution to the system \eqref{eq:pde} computed on a $128 \times 128$ grid, as described in Section \ref{sec:numerical_simulation}.
}
 \label{fig:example_dispersion_relation}
\end{figure}

We say $u^{\ast}$ is a \textit{Turing instability} of the system \eqref{eq:pde} if there exists a finite $\|q\|>0$ and some $i \in [N]$ for which $\text{Re}(\wt{\lambda}_i(q))>0$, i.e.~the steady state $u^*$ becomes unstable for a certain wavevector $q$. Here, we are interested in the additional condition $\text{Re}(\wt{\lambda}_j(q)) <0$ for $\|q\| \to \infty$ for \textit{all} $j\in [N]$, such that $\text{Re}(\wt{\lambda}_i(q))$ has a global maximum for some finite $\|q\|$. This type of instability typically leads to a \textit{stable} pattern of a wavelength corresponding to $q$ \cite{murray2001mathematical}. In summary, the conditions for a Turing instability read
\begin{subequations}
\begin{align}
    &
    \text{Re}(\wt{\lambda}_j(0)) = \text{Re}(\lambda_j) <0,\;\text{for all}\; j\in [N], \\
    &
    \text{there exist} \; \|q\|>0 \;\text{and}\; i \in [N] \;\text{for which}\;
    \text{Re}(\wt{\lambda}_{i}(q)) > 0, \\
    &
    \text{Re}(\wt{\lambda}_j(q)) <0 \;\text{if}\; \|q\| \to \infty\;\text{for all}\; j\in [N].
\end{align}
\end{subequations}
For other types of instabilities, we refer the reader to \cite{scholes2019comprehensive}.
Figure \ref{fig:example_dispersion_relation} shows $\text{Re}(\wt{\lambda}_i(q))$ of a two-species system ($N=2$) with Turing instability and the Turing pattern resulting from solving numerically the system of equations \eqref{eq:pde}. 

%

\subsection{Numerical Simulation}\label{sec:numerical_simulation}

This section describes the numerical algorithm used to simulate the system of PDEs \eqref{eq:pde}. 

\subsubsection{Discretization}\label{sec:discretization}
We consider reaction-diffusion systems of the form \eqref{eq:pde} with two spatial dimensions $M=2$, spatial points and domain 
\begin{equation}\label{eq:domain_size}
r=(r_1,r_2) \in S=[0, n_r] \times [0,n_r], \quad n_r \in \mathbb{N},
\end{equation}
and their solutions within the time interval $t\in [0,T]$. We further assume doubly periodic boundary conditions, i.e., $u_i(0, r_2,t) = u_i(n_r, r_2,t)$ for each $r_2 \in [0,n_r]$ and  $u_i(r_1, 0,t) = u_i(r_1, n_r,t)$ for each $r_1 \in [0,n_r]$, $t \in [0,T]$ and $i \in [N]$. The domain $S$ is discretized into a regular torus grid graph $G=(V,E)$ of size 
\begin{equation}\label{eq:def-m-V}
m = |V|=n_{r}\times n_{r},
\end{equation}
where each node $v\in V$ indexes a point $r_{v}\in S$. The edge set $E$ represents the adjacency of each node to its four nearest neighbors on the grid and takes into account the doubly periodic boundary conditions. Each of these edges have length $1$ corresponding to uniform sampling along each coordinate $r_{1}$ and $r_{2}$, respectively, of size $1$.

\subsubsection{Algorithm}
\textbf{Case $N=1$.} For simplicity, consider first the case of a single species $N=1$, $u(r,t) = u_1(r,t)$, with diffusion constant $\delta$, and let $v(t)$ denote the vector obtained by stacking the rows of the two-dimensional array of function values $\big(u(r_{v},t)\big)_{v\in V}$ evaluated on the grid. We discretize time into equal intervals of length $h> 0$ and write $v^{(k)} = v(k h)$ for $k \in \mathbb{N}$. The discretized PDE of the single species case 
\begin{equation}
    \frac{\partial}{\partial t}u(r,t) = f\big(u(r,t)\big) + d\Delta u(r,t),\quad u(r,0)=u_{0}(r)
\end{equation}
of Eq.~\eqref{eq:pde} is solved by the implicit Euler scheme
\begin{align}\label{eq:pde_discretized}
  \frac{v^{(k+1)} - v^{(k)}}{h} 
  & =
    f(v^{(k+1)}) + \delta L v^{(k+1)},
\end{align}
where matrix $L$ is the Laplacian discretized using the standard 5-point stencil. 
To perform a single time-step update according to \eqref{eq:pde_discretized}, we rewrite this equation as fixed point iteration with an inner iteration index $l$
\begin{align}\label{eq:fixed_point}
  v^{(k_l)} 
  & =
    (I - h \delta L)^{-1} \big(v^{(k)} + h f(v^{(k_{l-1})})\big), \quad l=1,2,\ldots, \quad v^{(k_0)}=v^{(k)},
\end{align}
where $I$ is the identity matrix. This fixed point equation is iterated until the convergence criterion
\begin{equation}\label{eq:inner_convergence_criterion}
    \frac{\|v^{(k_{l})}-v^{(k_{l-1})}\|}{\|v^{(k_{l-1})}\|} \leq \veps_{l}
\end{equation}
is met for some constant $0 < \veps_{l} \ll 1$, followed by updating the outer iteration \eqref{eq:fixed_point}
\begin{align}
  v^{(k+1)} = v^{(k_{l})}. 
\end{align}
The outer iteration is terminated when
\begin{equation}\label{eq:outer_convergence_criterion}
    \|f(v^{(k+1)}) + \delta L v^{(k+1)}\|_{\infty} \leq \veps_{k}
\end{equation}
for some constant $0 < \veps_{k} \ll 1$.

Due to the doubly periodic boundary conditions, the matrix $I - h \delta L = W \Lambda W^{\ast}$ is a sparse, \emph{block-circulant} and can hence be diagonalized using the unitary Fourier matrix $W$ corresponding to the two-dimensional discrete Fourier transform of doubly periodic signals defined on the graph $G$. As a result, using the fast Fourier transform (2D-FFT), multiplication of the inverse matrix by some vector $b$,
\begin{align}
  (I - h \delta L)^{-1}b
  & =
    W \Lambda^{-1} W^{\ast} b,
\end{align}
can be efficiently computed due to the convolution theorem by
\begin{itemize}
    \item computing the 2D-FFT of $b$: $\wh{b} = W^{\ast} b$,
    \item pointwise multiplication with the inverse eigenvalues of the matrix: $\Lambda^{-1} \hat{b}$, where $\Lambda^{-1}$ is a diagonal matrix and hence is inverted \textit{elementwise},
    \item transforming back using the inverse 2D-FFT: $W (\Lambda^{-1} \wh{b})$. 
\end{itemize}
The eigenvalues $\Lambda$ of the matrix $I+h \delta L$ result from applying the 2D-FFT to the block-circulant matrix corresponding to the convolution stencil
\begin{align}
  \begin{pmatrix}
    0 & 0 & 0 \\
    0 & 1 & 0 \\
    0 & 0 & 0
  \end{pmatrix}
  +
  h \delta
  \begin{pmatrix}
    0 & -1 & 0 \\
    -1 & 4 & -1 \\
    0 & -1 & 0
  \end{pmatrix}.
\end{align}

\textbf{Case $N>1$.} This procedure applies almost unchanged to the case of \textit{multiple species} ($N>1$), because the diffusion operator $D\Delta_{N}$ of \eqref{eq:pde} is block-diagonal. It suffices to check the case $N=2$: $v(t) = \bsm v_{1}(t) \\ v_{2}(t) \esm$ denotes the stacked subvectors $v_{1}, v_{2}$ that result from stacking the rows of the two-dimensional arrays of function values $\big(u_{1}(r_{v},t)\big)_{v\in V}, \big(u_{2}(r_{v},t)\big)_{v\in V}$ evaluated on the grid. The fixed point iteration \eqref{eq:fixed_point} then reads 
\begin{align}\label{eq:fixed_point-N=2}
  \bpm v_{1}^{(k_l)} \\ v_{2}^{(k_l)} \epm
  & =
    \bpm (I - h \delta_{1} L)^{-1} & 0 \\ 0 & (I - h \delta_{2} L)^{-1} \epm 
    \left(\bpm v_{1}^{(k)} \\ v_{2}^{(k)} \epm + 
    h \bpm f_{1}(v_{1}^{(k_{l-1})},v_{2}^{(k_{l-1})}) \\ f_{2}(v_{1}^{(k_{l-1})},v_{2}^{(k_{l-1})})\epm\right), 
\end{align}
where the mappings $(I-h \delta_{i}L)^{-1},\,i=1,2$ can be applied in parallel to the corresponding subvectors. Note that the vector $f(v^{(k_{l-1})}) = \bsm f_{1}(v^{(k_{l-1})}) \\ f_{2}(v^{(k_{l-1})}) \esm$ \textit{couples} the species concentrations. The general case $N>2$ is handled similarly.

\subsubsection{Step Size Selection}\label{sec:step-size-selection}
Step size $h$ has to be selected such that two conditions are fulfilled: Matrix $I-h \delta L_{N}$ should be invertible where $\delta L_{N}$ means the block-diagonal matrix 
\begin{equation}
d L_{N} = \Diag(d_{1} L,\dotsc,d_{N} L),
\end{equation}
and the fixed point iteration \eqref{eq:fixed_point} should converge. We discuss these two conditions in turn.

The first condition holds if $I-h \delta_{i} L$ is invertible for every $i\in [N]$, which certainly holds if $\lambda_{\min}(I-h \delta_{i} L) > 0$ which yields
\begin{equation}\label{eq:h-bound-1}
    h < \frac{1}{\max\{\delta_{i}\}_{i\in [N]} \lambda_{\max}(L)}.
\end{equation}
This also yields the estimate
\begin{equation}\label{eq:IhdLN-1-estimate}
    \|(I-h \delta L_{N})^{-1}\| = \frac{1}{\lambda_{\min}(I-h \delta L_{N})} \leq \frac{1}{1-h \max\{\delta_{i}\}_{i\in [N]} \lambda_{\max}(L)}.
\end{equation}
$\lambda_{\max}(L)$ may be easily computed beforehand using the power method \cite[p.~81]{Horn:2013aa} or replaced by the upper bound due to Gerschgorin's circle theorem \cite[Section 6.1]{Horn:2013aa}.

Now consider the fixed point iteration \eqref{eq:fixed_point}. 
Due to our assumptions stated after eq.~\eqref{eq:ode}, the mapping $f$ is Lipschitz continuous, i.e.~there exists a constant $L_{f}>0$ such that
\begin{equation}\label{eq:def-Lf}
    \|f(v)-f(v')\| \leq L_{f}\|v-v'\|,\qquad \forall v, v'.
\end{equation}
Thus, writing $T_{h}(v)=(I-h \delta L_{N})^{-1}\big(v^{(k)}-h f(v)\big)$, we obtain using \eqref{eq:IhdLN-1-estimate} and \eqref{eq:def-Lf}
\begin{equation}\label{eq:h-bound-2}
    \|T_{h}(v)-T_{h}(v')\| \leq \frac{h L_{f}}{1-h \max\{\delta_{i}\}_{i\in [N]} \lambda_{\max}(L)} \|v-v'\|,\quad \forall v,v'.
\end{equation}
As a result, both above-mentioned conditions hold if $h$ is chosen small enough to satisfy \eqref{eq:h-bound-1} and
\begin{equation}
    \frac{h L_{f}}{1-h \max\{\delta_{i}\}_{i\in [N]} \lambda_{\max}(L)} < 1.
\end{equation}
Then the mapping $T_{h}$ is a contraction and, by Banach's fixed point theorem, the iteration converges.

\subsection{The Gierer-Meinhardt model}\label{sec:examples}

As concrete examples, we consider evaluations of the Gierer-Meinhardt model \cite{gierer1972theory} comprising two species: a slowly diffusing activator that promotes its own and the other species' production, and a fast diffusing inhibitor that suppresses the production of the activator. Regarding the representation of the model by means of a PDE as in eq.~\eqref{eq:pde}, several different variants have been proposed in the literature \cite{gierer1972theory, murray2001mathematical}. Here, we use the dimensionless version analysed in \cite{murray1982parameter} and defined by
\begin{align}\label{eq:gm_model}
  f(u)
  & =
    \begin{pmatrix}
      a - b u_1 + \frac{u_1^2}{u_2(1 + cu_1^2)} \\
      u_1^2 - u_2
    \end{pmatrix}, \quad
  D 
  =
    s \begin{pmatrix}
      1 & 0 \\
      0 & \delta
    \end{pmatrix},
\end{align}
with parameters $a,b, c, \delta, s > 0$ and the shorthands (cf.~Eq.~\eqref{eq:pde})
\begin{equation}
u_1 = u_1(r,t),\qquad u_2 = u_2(r,t).
\end{equation}
Since only the ratio between the diffusion constants of the two species effects the stability of the system, the diffusion constant of the first species in \eqref{eq:gm_model} is normalised and we set $\delta_{1}=1, \delta_{2}=\delta$. The overall scaling of $D$ determines however the wavelength of an emerging pattern, and we accordingly multiply the diffusion matrix $D$ in eq.~\eqref{eq:gm_model} with an additional scaling factor $s > 0$.

\begin{figure*}[t]
\begin{center}
\centerline{\includegraphics[width=1.0\textwidth]{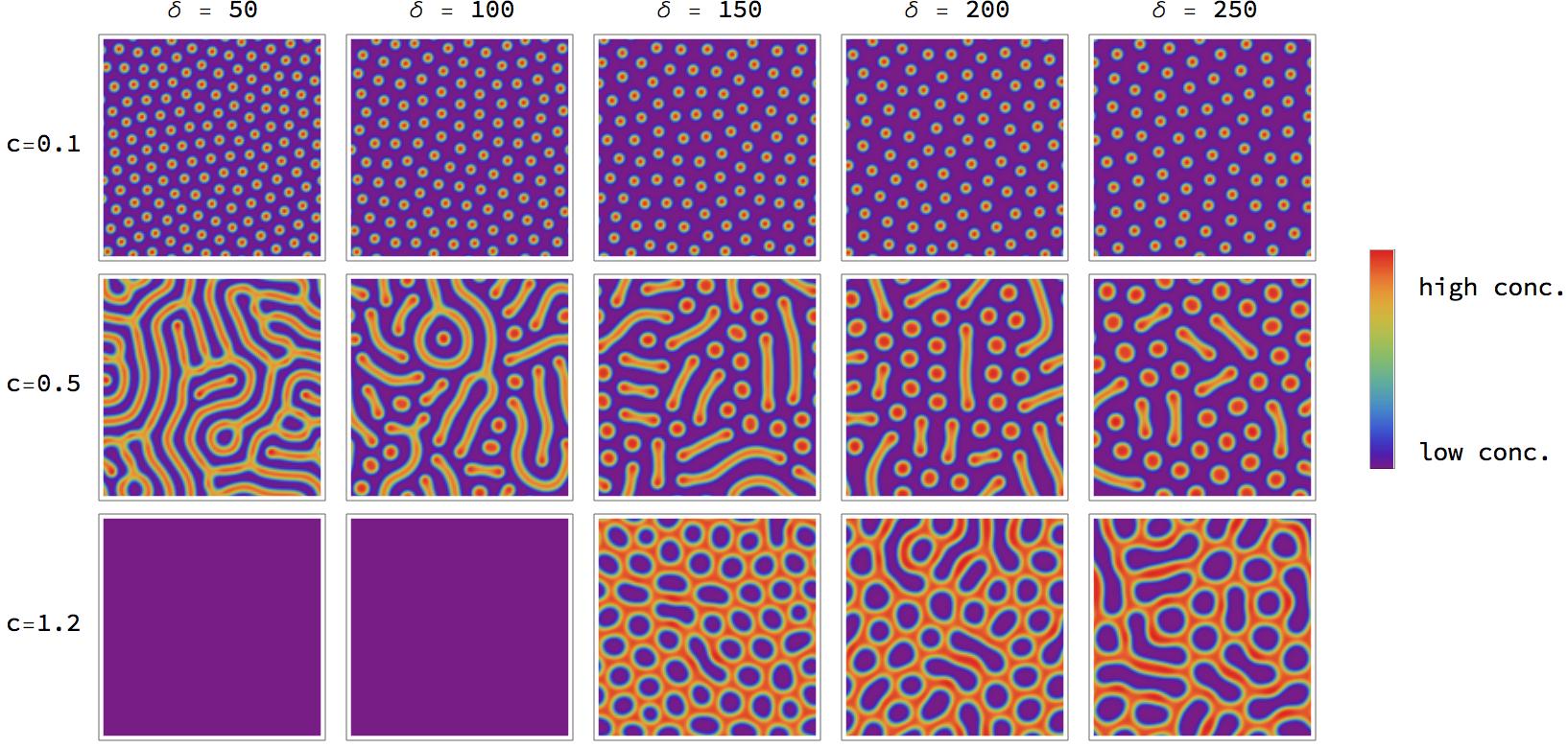}}
\caption{Simulation results of species $u_1$ in the Gierer-Meinhardt model defined by Eq.~\eqref{eq:gm_model} on a $128 \times 128$ grid with final time $T=5000$ for varying parameters $c$ and $\delta$ and fixed parameters $a=0.02$, $b=1.0$ and $s=0.5$. We observe that different parameter combinations give rise to different types of patterns and differing wavelengths. For $c=1.2$ and $\delta=50$ and $\delta=100$ we find a homogeneous solution and no pattern, which illustrates that the system does not exhibit an Turing instability for these parameter values.
 }\label{fig:gm_example_patterns}
\end{center}
\end{figure*}

Figure \ref{fig:example_dispersion_relation} displays the eigenvalues of the Jacobian of this model in the context of Turing instabilities as described in Section \ref{sec:Turing-patterns}, for one choice of parameters $a,b, c, \delta$, and the corresponding Turing pattern emerging when simulating the model numerically as described in Section \ref{sec:numerical_simulation}. Figure \ref{fig:gm_example_patterns} shows simulation results for various parameter values $c$ and $\delta$. The model gives rise to different types of patterns, ranging from spots to labyrinths. The characteristic length scale, or `wavelength', of the pattern varies with these parameters. This limits the ranges of parameter values that we analyse in the experiments: a too small wavelength leads to numerical artefacts when simulating the corresponding PDEs due to discretization errors; a too large wavelength on the other hand only yields a small section of the spatial pattern as `close-up view'. 

We hence only consider parameter values that exclude both extreme cases relative to a fixed grid graph that was used for numerical computation. 

%

\section{Extracting Features from Turing Patterns}\label{sec:Feature-extraction}
We extract two types of features from Turing patterns: \textit{resistance distance histograms} (Section \ref{sec:resistance_histograms}) efficiently encode the spatial structure of patterns due to their stability under spatial transformations. This almost invariant representation also includes few symmetries, however, 
which may reduce the accuracy of parameter prediction in certain scenarios. Hence two additional features are extracted that remove some of these symmetries (Section \ref{sec:additional_features}).

\subsection{Resistance Distance Histograms (RDHs)}\label{sec:resistance_histograms}

Resistance distance histograms (see Definition \ref{def:RDH} below) require two standard preprocessing steps described subsequently: representing Turing patterns as weighted graphs and computing pairwise resistance distances.

\paragraph{Graph-based representation of Turing patterns.} 

Let $u_{i,j}, i,j \in [n_r]$ be the concentration values of some species of a reaction-diffusion system at time $t=T$ on a regular torus grid graph $G=(V,E)$ of size $m=|V|=n_{r}\times n_{r}$, where each node $v\in V$ indexes a point $r_{v}\in S =[0, n_{r}] \times [0,n_{r}]$ (cf.~Section \ref{sec:discretization}). Let $u_v=u_{i,j}$ be the concentration value at $v=(i,j)$, obtained by simulating a system of PDEs \eqref{eq:pde} as described in Section \ref{sec:numerical_simulation}, and denote by
\begin{align}
  \ol{u} 
  & = 
    \frac{1}{m} \sum_{v \in V} u_v
\end{align}
the mean concentration.  We assign weights $\omega_{vv'}$ to edges $(v, v') \in E$ between adjacent nodes $v, v' \in V$ by
\begin{align}\label{eq:edge_weights}
  \omega_{vv'} 
  & =
    \begin{cases}
    1,  & \text{if}\; (u_v \geq \ol{u} \text{ and } u_{v'} \geq  \ol{u}) \text{ or } (u_v < \ol{u} \text{ and } u_{v'} <  \ol{u}), \\
    \epsilon, & \text{otherwise, where } 0 < \epsilon \ll 1,
    \end{cases}
\end{align}
that is edges between adjacent nodes receive the unit weight $1$ if both concentrations are either larger or smaller than the mean concentration $\ol{u}$, and the weight $\epsilon$ otherwise. 

\begin{figure*}[t]
\centerline{\includegraphics[width=0.9\textwidth]{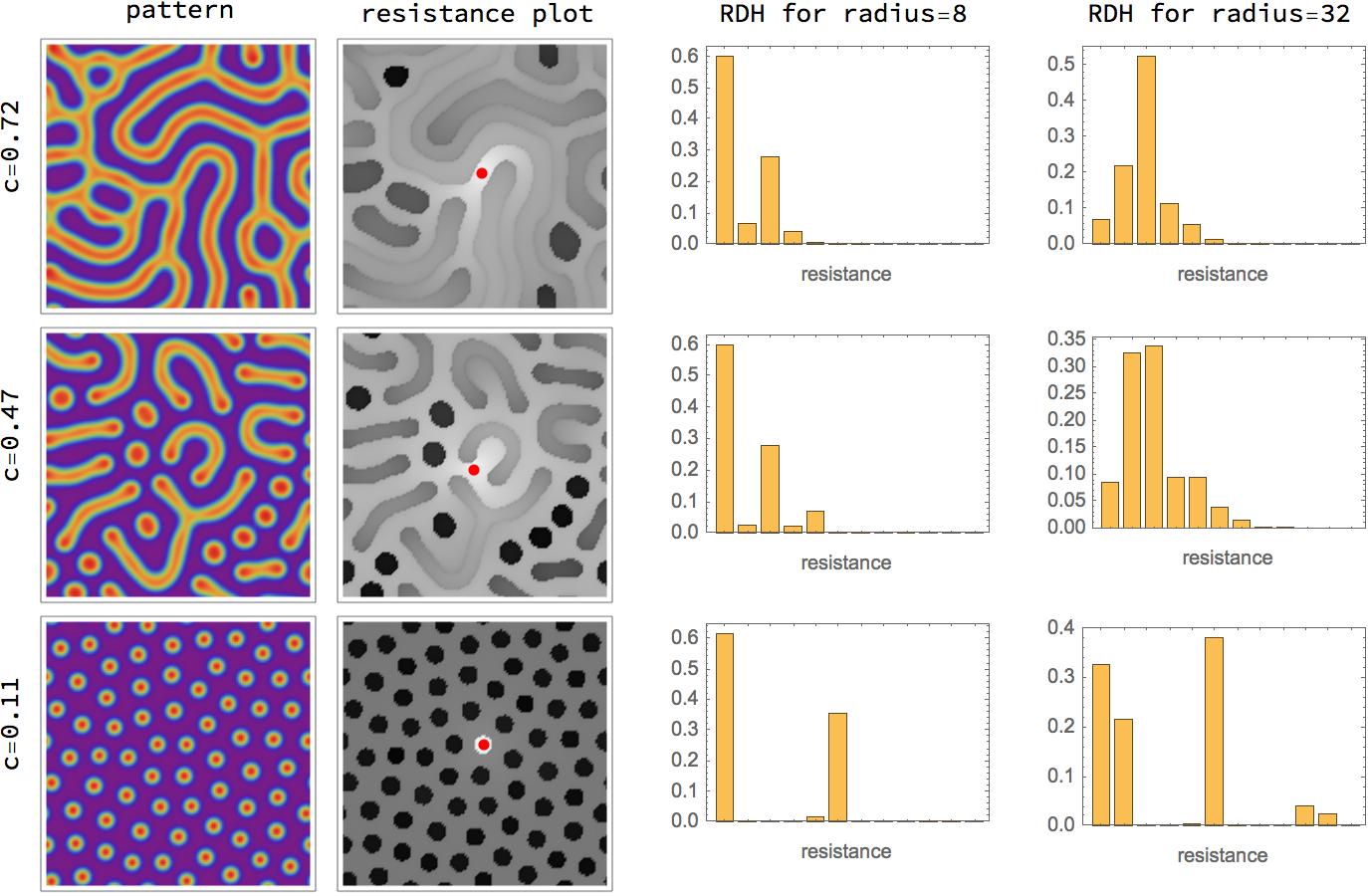}}
\caption{Each row of the figure shows from left to right: a pattern, a resistance distance plot and resistance distance histograms (RDH) for radii $8$ and $32$ of the first species, obtained from simulating the Gierer-Meinhardt model in Eq.~\eqref{eq:gm_model} on a $128 \times 128$ grid. Rows correspond to different values of parameter $c$. Weights are assigned to the edges of the corresponding grid graph according to Eq.~\eqref{eq:edge_weights}.
The resistance plots visualise resistance distances between all nodes and one central node marked with red. These plots result from partitioning the column of the resistance matrix $R$ \eqref{eq:def-Rvv'} corresponding to the central node into an $128 \times 128$ array. The darker the colour of a node the larger its resistance towards the central node. One can observe how the resistance values vary depending on the local structure of the pattern. In particular, the resistance distance histograms (RDHs) (cf.~Definition \ref{def:RDH}) differ substantially for the different types of patterns. The parameters are set to $a=0.02$, $b=1$, $\delta=100$ and $s=0.8$ for all three columns, and the parameter c is set to $0.72, 0.47$ and $0.11$ for the three columns, respectively. The final simulation time is $T=5000$. The RDHs are computed for $B=12$ bins and hypergraph spacing of $t=1$.
 }\label{fig:example_resistance_features}
\end{figure*}
%

\paragraph{Resistance distances and histograms.}
Based on \eqref{eq:edge_weights}, we define the weighted adjacency matrix $\Omega_G$ and graph Laplacian $L_G
$ of $G$,
\begin{align}
   \Omega_G 
   & = 
     (\omega_{vv'})_{v,v' \in V}, \\ 
  D_{G} &= \Diag(\Omega_G \mathbb{1}_m), \\ \label{eq:def-LG}
  L_G
  & = D_{G} - \Omega_{G},
\end{align}
where $\mathbb{1}_m$ is an $m$-dimensional column vector with all entries equal to $1$. Using \eqref{eq:def-LG}, in turn, we define the \emph{Gram} or \textit{kernel matrix} $K$ 
\begin{align}\label{eq:def-Gram-K}
  K
  & = 
    (J_m + L_G)^{-1} \in \mathbb{R}^{m \times m}, \qquad J_{m} = \mathbb{1}_{m} \mathbb{1}_{m}^{T} \in \mathbb{R}^{m \times m}
\end{align}
and the \emph{resistance matrix} $R \in \mathbb{R}^{m \times m}$ 
\begin{align}\label{eq:def-Rvv'}
  R
  & =
    (R_{vv'})_{v,v' \in V}, \quad
  R_{vv'} 
  = 
    K_{vv} +  K_{v'v'} - 2 K_{v'v}, \quad v,v' \in V.
\end{align}
Each entry $R_{vv'}$ is the \textit{resistance distance} between $v$ and $v'$ that was introduced in \cite{Klein:1993aa}. Its name refers to relations of the theory of electric networks \cite{Doyle:1984aa}, \cite[Chapter 10]{Bapat:2014aa}, \cite[Chapter 8]{Bremaud:2017aa}. A \textit{geometric interpretation} results from the relation 
\begin{equation}
    R_{vv'} \leq d_{G}(v,v'),
\end{equation}
where $d_{G}$ denotes the length of the shortest weighted path connecting $v$ and $v'$ in $G$. The bound is tight if this path is unique. Conversely, if multiple paths connect $v$ and $v'$, then the resistance distance is strictly smaller than $d_{G}(v,v')$. This sensitivity to the connectivity between nodes in graphs explains its widespread use, e.g.~for cluster and community detection in graphs \cite{Fortunato:2010aa}.

A \textit{probabilistic interpretation} of the resistance distance is as follows. Consider a random walk on $G$ performing jumps along the edges in discrete time steps, and assume that the probability to jump along an edge is proportional to the edge's weight. Then $R_{vv'}$ is inversely proportional to the probability that the random walk starting at $v$ visits $v'$ before returning to $v$ \cite[Section 10.3]{Bapat:2014aa}. In view of \eqref{eq:edge_weights}, this implies that the process jumps more likely between neighbouring nodes with both large (small) concentrations than between differing concentrations. 

We add a third interpretation of the resistance distance from the viewpoint of kernel methods \cite{Hofmann:2008aa,Seto:2014aa} and \textit{data embedding}. Let
\begin{equation}
    \mc{F}_{G} = \{f \colon V \to \R\} \cong \R^{m}
\end{equation}
the space of functions on $V$ that we identify with real vectors of dimension $m=|V|$, and consider the bilinear form
\begin{equation}
    \mc{E}\colon\mc{F}_{G}\times\mc{F}_{G} \to \R,\qquad
    \mc{E}(f,g)=\frac{1}{2}\sum_{v,v'\in V}\w_{vv'}(f_{v}-f_{v'})(g_{v}-g_{v'}) = \la f, L_{G} g\ra.
\end{equation}
Since $G$ is connected, the symmetric and positive semidefinite graph Laplacian $L_{G}$ has a single eigenvalue $0$ corresponding to the eigenvector $\eins_{m}$. Consequently, using $\mc{E}$, one defines the Hilbert space 
\begin{subequations}
\begin{align}
\mc{H}_{G} &=(\mc{F}_{G},\la\cdot,\cdot\ra_{G})
\intertext{with inner product}\label{eq:def-ip-G}
    \langle f,g\rangle_{G} &= \Big(\sum_{v\in V} f_{v}\Big)\Big(\sum_{v'\in V} g_{v'}\Big) + \mc{E}(f,g)
    = \big\la f, (J_{m}+L_{G}) g\big\ra.
\end{align}
\end{subequations}
Since $\dim\mc{H}_{G}<\infty$, all norms are equivalent and the evaluation map $f \mapsto f_{v}$ is continuous. Hence $\mc{H}_{G}$ is a \textit{reproducing kernel Hilbert space} \cite[Def.~1.1]{Paulsen:2016aa} with reproducing kernel
\begin{equation}
    K \colon V \to V \to \mc{H}_{G},\qquad
    K(v,v') = K_{vv'} = \la K^{v},K^{v'}\ra_{G},\quad v,v' \in V,
\end{equation}
where $K_{vv'}$ denotes the entries of the Gram matrix \eqref{eq:def-Gram-K}, and $K^{v},K^{v'}$ are the column vectors indexed by $v, v'$ and interpreted as elements (functions) in $\mc{H}_{G}$. The resistance distance \eqref{eq:def-Rvv'} then takes the form
\begin{equation}
    R_{vv'} = \|K^{v}-K^{v'}\|_{G}^{2},\qquad v,v'\in V,
\end{equation}
where $\|\cdot\|_{G}$ denotes the norm induced by the inner product \eqref{eq:def-ip-G}. This makes explicit the \textit{nonlocal} nature of the resistance distance $R_{vv'}$ between nodes $v,v'\in V$ in terms of the squared distance of the corresponding functions $K^{v},K^{v'}$ in $\mc{H}_{G}$. 

Overall, each of the three interpretations reveals how the resistance distance measures \textit{nonlocal spatial structure} of Turing patterns. 
Figure \ref{fig:example_resistance_features}(b) visualises the resistance distances $\{R_{vv'}\}_{\,v'\in V}$ with respect to one fixed node $v$. We condense these data extracted from Turing patterns into features, in terms of corresponding histograms described next.
\begin{definition}[resistance distance histogram (RDH)] \label{def:RDH}
Let $V_{t} \subset V,\, t \in \N$ denote the nodes of the subgraph induced by the subgrid  that is obtained from the underlying grid graph $G=(V,E)$ by undersampling each coordinate direction with a factor $t$. Define the set of resistance distance values
\begin{equation}\label{eq:def-mcR}
    \mc{R}_{t,r} = \{R_{vv'}\colon v\in V_{t},\, v'\in V,\,\|r_{v}-r_{v'}\|\leq r\}
\end{equation}
parametrized by $t$ and a radius parameter $r\in\R$. The resistance distance histogram (RDH) $H_{r,t} \in \Delta_{B}$ is the normalized histogram of the resistance distance values $\mc{R}_{t,r}$ with respect to a uniform binning with bin number $B$ of the interval $[0,R_{\max}]$, with a suitably chosen maximal value $R_{\max}$.
\end{definition}
The radius parameter $r$ specifies the spatial scale at which the local structure of Turing patterns is assessed through RDHs, in a way that is stable against spatial transformations of the local domains corresponding to \eqref{eq:def-mcR}. $r$ is the only essential parameter, since RDHs are based on data $\mc{R}_{r,t}$ collected from nodes $v\in V_{t}$. This results in averaging of local pattern structure and makes RDHs only weakly dependent on the spacing $t$. In addition, the representation becomes robust against local noise in the pattern caused by random initial conditions.

\begin{figure*}[t]
\begin{center}
\centerline{
\parbox{0.3\textwidth}{\centering 
\includegraphics[width= \linewidth]{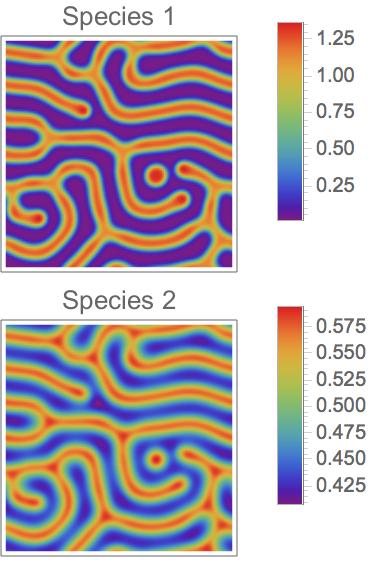}
(a)
}\hfill
\parbox{0.3\textwidth}{\centering 
\includegraphics[width= \linewidth]{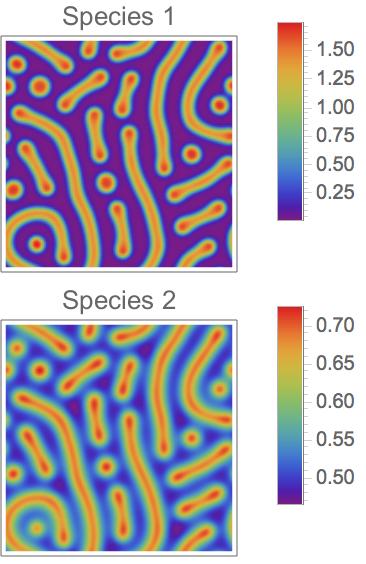}
(b)
}\hfill
\parbox{0.3\textwidth}{\centering 
\includegraphics[width= \linewidth]{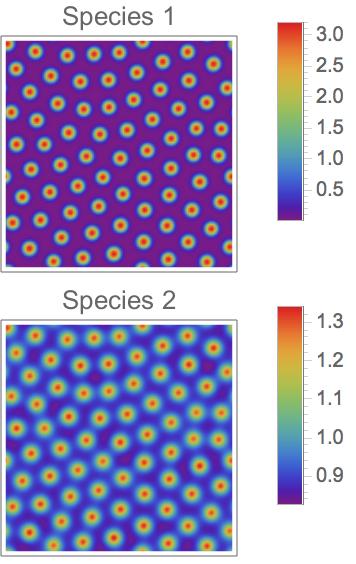}
(c)
}
}
\caption{The patterns depicted by Figure \ref{fig:example_resistance_features} are shown  again, here for \textit{both} species, however. (a), (b) and (c) correspond to the three rows of Figure \ref{fig:example_resistance_features}. It is apparent that the patterns of the two species are qualitatively very similar: Basically, they are just rescaled and shifted versions of each other. Since resistance distance histograms (RDHs) are invariant under such operations, the resulting RDHs would be approximately equal. It is hence sufficient to use only the patterns of one species for computing RDHs and subsequent analysis. 
 }\label{fig:gm_example_patterns_two_species}
\end{center}
\end{figure*}

\begin{remark}\label{rem:redundancy_different_species}
Typically, concentrations of different species in a Turing pattern are  approximately scaled (and sometimes reflected) versions of each other, in particular for two-species systems like the Gierer-Meinhardt model studied here. See Figure \ref{fig:gm_example_patterns_two_species} for an illustration. The RDHs defined in Definition \ref{def:RDH} hence contain redundant information when computed for different species. Therefore, we only use concentrations of one species to compute RDHs in the following.
\end{remark}

\subsection{Maximal Concentration and Connected Components}\label{sec:additional_features}

RDHs according to Definition \ref{def:RDH} represent the spatial structure of Turing patterns in a compact way. However, 
this representation also includes few symmetries such that certain properties of patterns are not captured, such as
\begin{enumerate}
    \item \textit{absolute concentration values}: rescaling or shifting the concentration values of a pattern does not change the RDHs;  
    \item \textit{range-reflection symmetry}: reflection of a pattern on any plane of constant concentration, i.e.~inverting the total order that defines the weights \eqref{eq:edge_weights}, does not change the RDHs. 
\end{enumerate}
To account for these two properties, we introduce the following two additional features.
\begin{enumerate}
    \item \textbf{Maximal concentration $c_m$}: We aim to estimate the concentration of areas in the pattern with large concentrations while disregarding local fluctuations that potentially arise from numerical inaccuracies.  To this end, we bin the concentration values of a pattern into a histogram and define the maximal concentration $c_m$ as the location of the right-most peak. 
    \item \textbf{Number of connected components $n_c$}: To account for the above-mentioned range reflection symmetry, we define the graph $G' = (V,E')$ with the same nodes $V$ as the original graph $G$ but with only a subset of edges $E' \subset E$ between nodes of high concentration. We then compute the number $n_c$ of connected components in $G'$.
\end{enumerate}
This list of pattern properties not captured by RDHs is not exhaustive, of course. For example, RDHs do not effectively measure the steepness of transitions between areas of high to low concentrations. However, since RDHs turned out to be powerful enough for parameter estimation, as 
shown below, we did not use further features in this study.

\section{Learning Parameters from Spatial Turing Patterns}\label{sec:Learning-Patterns}

This section concerns the problem of learning the kinetic parameter of Turing patterns described in Section \ref{sec:Turing-patterns}. We start by formulating the learning problem in Section \ref{sec:learning_problem}. Sections \ref{sec:svr} and \ref{sec:neural_networks} introduce the approaches for parameter prediction studied in this paper, kernel based predictors and neural networks, respectively.

\subsection{Setup}\label{sec:learning_problem}

\subsubsection{Learning Problem}\label{sec:sub_learning_problem}

We consider a multi-output learning problem with training set 
\begin{equation}\label{eq:def-mcDn}
\mc{D}_{n} = \{(x_i, y_i)\}_{i\in [n]} \subset \mc{X}\times\mc{Y},\qquad n\in \mathbb{N},\quad \mc{X}=\{ H_{r,t}\},\quad \mathcal{Y} = \mathbb{R}^d,
\end{equation}
where each $x_i$ is a resistance distance histogram $H_{r,t}$ (RDH) according to Definition \ref{def:RDH}, 
and vectors $y_i$ comprise parameter values of a model, such as the parameters $a,b,c,\delta$ of the Gierer-Meinhardt model \eqref{eq:gm_model}. 
Our goal is to learn a prediction function $f:\mathcal{X} \to \mathcal{Y}$ that generalises well to points $(x,y) \notin \mc{D}_{n}$.

We distinguish \textit{individual} parameter prediction (Section \ref{sec:svr-separable}) corresponding to dimension $d=1$, where for \textit{each} model parameter a predictor function $f$ is \textit{separately} learned, and \textit{joint} parameter prediction (Section \ref{sec:operator-valued_kernel}) corresponding to $d>1$, where a \textit{single vector-valued} predictor function $f$ is learned. In each of these cases, the output training data $y_{i}$ in $\mc{D}_{n}$ \eqref{eq:def-mcDn} have to be interpreted accordingly.

Regarding individual parameter prediction, we employ basic support vector regression \cite{Evgeniou:2000vo,Smola:2004aa} in Section \ref{sec:svr-separable} and specify suitable kernel functions for resistance histograms as input data in Section \ref{sec:kernel-functions}. Regarding joint parameter prediction, we employ regression using an operator-valued kernel function in Section \ref{sec:operator-valued_kernel}.  For background reading concerning reproducing kernel Hilbert spaces (RKHS) and their use in machine learning, we refer to \cite{Berg:1984aa,Berlinet:2004aa,Paulsen:2016aa} and \cite{Evgeniou:2000vo,cucker2001mathematical,Hofmann:2008aa}, respectively. We also utilize neural networks in Section \ref{sec:neural_networks} for both individual and joint parameter prediction.

\subsubsection{Accuracy Measure}\label{sec:accuracy_measure}

A commonly used measure for the accuracy of an estimator $f:\mathcal{X} \to \mathcal{Y}$ on a test set $X \times Y = \{(x_i, y_i)\}_{i\in [m]}, m\in \mathbb{N}, X \subset \mathcal{X} \subset \mathbb{R}^k, Y \subset \mathcal{Y} = \mathbb{R}^d$
is the \emph{root-mean-square error} (RMSE) defined as
\begin{align}\label{eq:rsme}
  \text{RMSE}
  & = 
  \Big(\frac{1}{m d}\ \sum_{i \in [m]} \|f(x_{i})-y_{i}\|^{2}\Big)^{1/2}
    = \Big(\frac{1}{m d}\ \sum_{i \in [m]} \sum_{j \in [d]} ((f_j(x_i) - y_{i,j})^2 \Big)^{1/2}.
\end{align}
Additional normalisation by the empirical mean value of the nonnegative target variables yields the \emph{normalised root-mean-square error} (NRSME) 
\begin{align}\label{eq:nrsme}
  \text{NRMSE}
  & = 
    \frac{\text{RMSE}}{\frac{1}{m d}\ \sum_{i \in [m]} \sum_{j \in [d]} y_{i,j}}.
\end{align}
We use the NRMSE to measure the accuracy of predicted parameter values in this work. Figure \ref{fig:patterns_for_varying_nrmse_values} (page \pageref{fig:patterns_for_varying_nrmse_values}) illustrates visually the variation of patterns for various NRMSE values. 

We consider as ``good'' model parameter predictions with accuracy values NRMSE $\leq 0.2$ and as ``excellent'' predictions with accuracy values NRMSE $\leq 0.05$.

\subsection{Kernel-Based Parameter Prediction}\label{sec:svr}

\subsubsection{Individual Parameter Prediction Using SVMs}\label{sec:svr-separable}

In this section, we focus on the case $d=1$ where along with a finite sample of RDHs $x_{i}$, values $y_{i}\in\R$ of some model parameter are given as training set \eqref{eq:def-mcDn}. \textit{Individual} parameter prediction means that, for each model parameter, an individual prediction function 
\begin{equation}\label{eq:f-individual}
f\colon\mc{X}\to\R
\end{equation}
specific to this particular parameter is determined. We apply standard support vector regression.

Given a symmetric and nonnegative kernel function (see Section \ref{sec:kernel-functions} for concrete examples)
\begin{equation}\label{eq:def-kernel-function}
  k\colon \mathcal{X} \times \mathcal{X} \to \R_{+},
\end{equation}
the corresponding \emph{reproducing kernel Hilbert space} (RKHS) with inner product $\la\cdot,\cdot\ra_{k}$ is denoted by $\mc{H}_{k}$. `Reproducing' refers to the property
\begin{subequations}\label{eq:properties-Hk}
\begin{align}
  k_{x} &= k(x,\cdot) \in \mc{H}_{k}, \quad \forall x \in \mc{X}, \\
  f(x) &= \langle f, k_{x} \ra_{k}, \quad \forall x \in \mathcal{X}, \quad \forall f \in \mathcal{H}_{k}.
\end{align}
\end{subequations}
Using the training set $\mc{D}_{n}$ and a corresponding loss function, our objective is to determine a prediction function $f\in\mc{H}_{k}$ that maps a RDH $x$ extracted from an observed pattern to a corresponding parameter value $y=f(x)$. We employ the $\veps$-insensitive loss function \cite{Evgeniou:2000vo}
\begin{equation}\label{eq:svr_loss}
\ell_{\veps}\colon\R\to\R_{+},\qquad
z\mapsto\ell_{\veps}(z)=\max\{0,|z|-\veps\},\qquad\veps\geq 0,
\end{equation}
to define the training objective function
\begin{equation}\label{eq:training-objective-separable}
   \sum_{i\in[n]}\ell_{\veps}\big(y_{i}-f(x_{i})\big) + \frac{\lambda}{2}\|f\|_{k}^{2},
\end{equation}
where the regularizing parameter controls the size of the set of prediction functions $f$ in order to avoid overfitting. Since $\ell_{\veps}$ is continuous and the regularizing term  monotonically increases with $\|f\|_{k}$, the representer theorem \cite{Whaba:1990uc,Scholkopf:2001aa} applies and implies that the function $f^{\ast}$ minimizing \eqref{eq:training-objective-separable} lies in the span of the functions generated by the training set
\begin{equation}\label{eq:f-ast-separable}
    f^{\ast} \in \Span\{k_{x_{1}},\dotsc,k_{x_{n}}\},\qquad 
    f^{\ast} = \sum_{i\in[n]}\alpha_{i}^{\ast}k_{x_{i}}.
\end{equation}
Using nonnegative slack variables $\xi_{i},\,i\in [n]$ in order to represent the piecewise linear summands $\ell_{\veps}\big(y_{i}-f(x_{i})\big)$ of \eqref{eq:training-objective-separable} by
\begin{equation}
\min_{\R}\xi_{i}\quad\text{subject to}\quad
\xi_{i} \geq 0,\quad
\xi_{i} \geq y_{i}-f(x_{i})-\veps,\quad
\xi_{i} \geq f(x_{i})-y_{i}-\veps,
\end{equation}
and substituting $f=\sum_{i\in[n]}\alpha_{i}k_{x_{i}}$ yields the training objective function  \eqref{eq:training-objective-separable} in the form
\begin{subequations}
\begin{align}\label{eq:svr_training_problem}
\min_{\alpha,\xi}\Big\{
\sum_{i} \xi_{i} + &
\frac{\lambda}{2}\la\alpha, K_{n}\alpha\ra\Big\},\quad
\lambda > 0,\qquad
K_{n} = \big(k(x_{i},x_{j})\big)_{i,j\in[n]}
\\
\text{subject to}\qquad
\xi_{i} &\geq 0, \\
\xi_{i} &+ \sum_{j}\alpha_{j}k(x_{j},x_{i})
\geq y_{i}-\veps, \\
\xi_{i} &- \sum_{j}\alpha_{j}k(x_{j},x_{i})
\geq -y_{i}-\veps.
\end{align}
\end{subequations}
Since $K_{n}$ is positive definite, this is a convex quadratic program that can be solved using standard methods. Substituting the minimizing vector $\alpha^{\ast}$ into \eqref{eq:f-ast-separable} determines the desired prediction function $f^{\ast}$.

This procedure is repeated to obtain prediction functions $f^{\ast}_{j},\,j\in[d]$ for each parameter to be predicted, using the training sets $\{(x_{i},y_{i,j})\}_{i\in[n]}$ for $j\in[d]$.

\subsubsection{Joint Parameter Prediction Using Operator-Valued Kernels}\label{sec:operator-valued_kernel}
In this section, we consider the general case $d \geq 2$: each vector $y_{i}$ of the training set \eqref{eq:def-mcDn} comprises the values of a fixed set of model parameters and $x_{i}$ is the RDH extracted from the corresponding training pattern. Our aim is to exploit dependencies between input variables $\{x_{i}\}_{i\in [n]}$ and output variables $\{y_{i}\}_{i\in [n]}$ as well as dependencies between the output variables. To this end, we use the method proposed by \cite{Kadri:2013vm} for vector-valued parameter prediction that generalizes vector-valued kernel ridge regression as introduced by \cite{Evgeniou:2005aa,Micchelli:2005aa}. See \cite{Alvarez:2012aa} for a review of vector-valued kernels and \cite{Brouard:2016aa,Minh:2016aa} for general operator-valued kernels and prediction.

In addition to a kernel function $k$ \eqref{eq:def-kernel-function} for the input data, we additionally employ a kernel function
\begin{equation}\label{eq:kernel-output}
    l\colon\mc{Y}\times\mc{Y}\to\R_{+},\qquad \mc{Y}=\R_{+}^{d}
\end{equation}
for the output data with corresponding RKHS $(\mc{H}_{l},\la\cdot,\cdot\ra_{l})$ and properties analogous to \eqref{eq:properties-Hk}, and a operator-valued kernel function
\begin{equation}
    K\colon\mc{X}\times\mc{X}\to \mc{L}(\mc{H}_{l})
\end{equation}
that takes values in the space $\mc{L}(\mc{H}_{l})$ of bounded self-adjoint operators from $\mc{H}_{l}$ to $\mc{H}_{l}$. $K$ enjoys the same properties as the more common scalar-valued kernel functions $k, l$, viz., it is the reproducing kernel of a RKHS $(\mc{H}_{K},\la\cdot,\cdot\ra_{K})$ of $\mc{H}_{l}$-valued functions
\begin{equation}
    g\colon\mc{X}\to\mc{H}_{l}.
\end{equation}
The properties analogous to \eqref{eq:properties-Hk} now read
\begin{subequations}\label{eq:K-RKHS-properties}
\begin{align}
    K(x,\cdot)\vphi &\in \mc{H}_{K}, &
    \forall x &\in\mc{X},\;\forall\vphi\in\mc{H}_{l},
    \\
    \la g,K(x,\cdot)\vphi\ra_{K} &= \la g(x),\vphi\ra_{l}, &
    \forall g &\in\mc{H}_{K},\;\forall x\in\mc{X},\;\forall\vphi\in\mc{H}_{l}.
\end{align}
\end{subequations}
In particular, the operator-valued kernel matrix
\begin{equation}
    K_{n}^{o}=\big(K(x_{i},x_{j})\big)_{i,j\in[n]}
\end{equation}
is positive definite, due to the positive definiteness of the kernel function $K$,
\begin{equation}
    \sum_{i,j\in[m]}\la \vphi_{i}, K(x_{i},x_{j})\vphi_{j}\ra_{l} \geq 0
\end{equation}
for all $m\in\N$, $x_{1},\dotsc,x_{m}\in\mc{X}$,  $\vphi_{1},\dotsc,\vphi_{m}\in\mc{H}_{l}$. 

In order to capture dependencies among the output variables as well as between input and output variables, the prediction function 
\begin{equation}\label{eq:f-vector-valued}
    f\colon\mc{X}\to\mc{Y}
\end{equation}
is not learned directly, unlike the individual predictors \eqref{eq:f-individual} in the preceding section (cf.~\eqref{eq:f-ast-separable}). Rather, the optimal mapping \eqref{eq:f-vector-valued} is parametrized by
\begin{equation}\label{eq:fast-vector}
    x\mapsto f^{\ast}(x) = \phi_{l}^{-1}\circ g^{\ast}(x),
\end{equation}
where $\phi_{l}\colon \mc{Y}\to\mc{H}_{l}$ is the feature map corresponding to the output kernel function \eqref{eq:kernel-output} (see, e.g., \cite[Section 3]{cucker2001mathematical}) satisfying 
\begin{equation}
\la\phi(y_{i}),\phi(y_{j})\ra_{l}=l(y_{i},y_{j}), 
\end{equation}
and $g^{\ast}\in\mc{H}_{k}$ is determined by regularized least-squares on the transformed training data $(x_{i},\phi_{l}(y_{i}))_{i\in[n]}$, that is by solving
\begin{equation}\label{eq:g-ast-problem}
    g^{\ast} = \arg\min_{g\in\mc{H}_{K}}\Big\{\sum_{i\in[n]}\|g(x_{i})-\phi_{l}(y_{i})\|_{l}^{2} + \lambda\|g\|_{K}^{2}\Big\},\quad \lambda > 0.
\end{equation}
Invoking again the representer theorem valid for the present more general scenario \cite{Micchelli:2005aa}, $g^{\ast}$ admits the representation
\begin{equation}\label{eq:g-ast-joint-feature}
    g^{\ast} = \sum_{i\in[n]}K_{x_{i}}\psi_{i}^{\ast},\quad
    \psi_{i}^{\ast}\in\mc{H}_{l},\quad K_{x_{i}}=K(x_{i},\cdot),
\end{equation}
which makes explicit how the approach generalizes the individual parameter predictors \eqref{eq:f-ast-separable}. 

It remains to specify a kernel function $K$ and the computation of $\phi_{l}^{-1}$ in order to evaluate the prediction map \eqref{eq:fast-vector}. As for $K$, our choice is
\begin{subequations}\label{eq:def-K-joint-feature}
\begin{align}
    K(x_{i},x_{j}) &= k(x_{i},x_{j})C_{\mc{YY}|\mc{X}}
\intertext{with the input kernel function $k$ \eqref{eq:def-kernel-function} and the conditional covariance operator 
}\label{eq:def-C-YYX}
C_{\mc{YY}|\mc{X}} &= C_{\mc{YY}} - C_{\mc{YX}}C_{\mc{XX}}^{-1}C_{\mc{XY}}
\end{align}
\end{subequations}
on $\mc{H}_{l}$. 
Since $K$ is evaluated on the training data, $C_{\mc{YY}|\mc{X}}$ is replaced in practice by evaluating the empirical covariance operators on the right-hand side, i.e.
\begin{equation}
    C_{n;\mc{YY}} = \frac{1}{n}\sum_{i\in[n]}l_{y_{i}}\otimes l_{y_{i}},
\end{equation}
with output kernel function $l$ \eqref{eq:kernel-output}, $l_{y_{i}}=l(y_{i},\cdot)$ and $(l_{y_{i}}\otimes l_{y_{j}})\vphi = \la l_{y_{j}},\vphi\ra_{l}l_{y_{i}}$, and similarly for the remaining mappings on the right-hand side of \eqref{eq:def-C-YYX}. The kernel function \eqref{eq:def-K-joint-feature} together with the predictor \eqref{eq:g-ast-joint-feature} in the \textit{output feature space} reveals how the dependencies are taken into account of both the output variables and between the input and output variables.

In order to obtain for some test input (RDH) $x$ the predicted parameter vector 
\begin{equation}
\wh{y}=f^{\ast}(x) = \phi_{l}^{-1}\circ g^{\ast}(x)
\end{equation}
from the predicted \textit{embedded} output value $g^{\ast}(x)$, the mapping $\phi_{l}^{-1}$ of \eqref{eq:fast-vector} has to be evaluated. This is an instance of the so-called \textit{pre-image problem} \cite{Scholkopf:1999aa,Honeine:2011aa}. In the present scenario, putting together \eqref{eq:g-ast-problem}, \eqref{eq:g-ast-joint-feature}, \eqref{eq:def-K-joint-feature} and \eqref{eq:K-RKHS-properties}, this yields after a lengthy computation \cite[Appendix]{Kadri:2013vm} the optimization problem
\begin{subequations}\label{eq:preimage-problem}
\begin{align}\label{eq:preimage-problem-a}
    \wh{y} &= \arg\min_{y\in\mc{Y}}\big\{l(y,y)-2 l_{y}^{\T}v(x;\mc{D}_{n})\big\} 
    \intertext{where}\label{eq:preimage-problem-v}
    v(x;\mc{D}_{n}) &= (k_{x}^{\T}\otimes T_{n})(K_{n}\otimes T_{n} + n\lambda I_{n^{2}})^{-1}\vvec(I_{n}),\quad\lambda > 0, 
    \\ \label{eq:preimage-problem-Tn}
    T_{n} &= L_{n}-(K_{n}+n\veps I_{n})^{-1} K_{n}L_{n},\quad 0 <\veps\ll 1, 
    \\ \label{eq:preimage-problem-Kn}
    K_{n} &= \big(k(x_{i},x_{j})\big)_{i,j\in[n]}, \qquad
    k_{x} = \big(k(x,x_{1}),\dotsc,k(x,x_{n})\big)^{\T}, 
    \\ \label{eq:preimage-problem-Ln}
    L_{n} &= \big(l(y_{i},y_{j})\big)_{i,j\in[n]}, \qquad\;\;
    l_{y} = \big(l(y,y_{1},\dotsc,l(y,y_{n})\big)^{\T}. 
\end{align}
\end{subequations}
Here, $\lambda$ in \eqref{eq:preimage-problem-v} is the regularization parameter of \eqref{eq:g-ast-problem}, $\veps$ in \eqref{eq:preimage-problem-Tn} is a small constant regularizing the numerical matrix inversion, $K_{n}, L_{n}$ are the input and output kernel matrices corresponding to the training data \eqref{eq:def-mcDn}, $x$ is a novel unseen test pattern represented as described in Section \ref{sec:Feature-extraction}, and $y$ is the parameter vector variable to be optimized.

Unlike the input kernel function $k$ that is applied to RDHs (see Section \ref{sec:kernel-functions}), the output kernel function $l$ applies to the common case of parameter vectors and hence choosing the smooth Gaussian kernel function as $l$ is a sensible choice. Therefore, once the vector $v(x;\mc{D}_{n})$ has been computed for a test pattern $x$, the optimization problem \eqref{eq:preimage-problem-a} can be solved numerically by iterative gradient descent with adaptive step size selection by line search.

Regarding the computation of the vector \eqref{eq:preimage-problem-v} that defines the objective function of \eqref{eq:preimage-problem-a}, the matrix $T_{n}$ given by \eqref{eq:preimage-problem-Tn} can be directly computed for numbers $n$ up to few thousands data points using off-the-shelf solvers. This is not the case for the linear system of \eqref{eq:preimage-problem-v} involving the Kronecker product $K_{n}\otimes T_{n}$, however, which is dense and has the size $n^{2}\times n^{2}$. Therefore, we solve the linear system
\begin{equation}\label{eq:Kronecker-system}
    (K_{n}\otimes T_{n}+n\lambda I_{n^{2}}) u = \vvec(I_{n})
\end{equation}
in a memory-efficient way using the global-GMRES algorithm proposed by \cite{Bouhamidi:2008vk} that iteratively constructs Krylov matrix subspaces and approximates the solution by solving a sequence of low-dimensional least-squares problems. Having computed $u$, the vector \eqref{eq:preimage-problem-v} results from computing
\begin{equation}
    v(x;\mc{D}_{n}) = \vvec(T_{v}\vvec^{-1}(u)k_{x}).
\end{equation}

\subsubsection{Kernels for Resistance Distance Histograms}\label{sec:kernel-functions}

In this section, we specify kernel functions \eqref{eq:def-kernel-function} that we evaluated for parameter prediction. Below, $x, x' \in \mc{H}_{r,t}$ denote two RDHs. 
\begin{description}
    \item[Symmetric $\chi^{2}$-kernel.] This kernel is member of a family of kernels generated by Hilbertian metrics on the space of probability measures on $\mc{X}$ \cite{Hein:2005aa} and defined by
\begin{align}\label{eq:chi_squared_kernel}
  k_{\gamma}(x,x')
  & =
     \sum_{i \in [B]} \frac{x_i x_i'}{x_i + x_i'}.
\end{align}
\item[Exponential $\chi^{2}$-kernel.]
The exponential $\chi^{2}$-kernel reads
\begin{align}\label{eq:chi_squared_kernel_exp}
  k_{\gamma}(x,x')
  & =
    \exp \Big( - \frac{1}{\gamma} \sum_{i \in [B]} \frac{(x_i - x_i')^2}{x_i + x_i'}  \Big),\quad\gamma > 0.
\end{align}

    \item[Wasserstein kernel.]
 We define a cost matrix 
\begin{equation}
C=(C_{i,j})_{i,j\in[B]},\qquad C_{i,j} = (i-j)^{2},\qquad i,j \in [B]
\end{equation}
and the squared discrete Wasserstein distance between $x$ and $x'$
\begin{equation}\label{eq:dW2}
d^{2}_{W}(x,x') = \la C, M^{\ast}\ra,
\end{equation}
where $M^{\ast}$ solves the discrete optimal transport problem \cite{peyre2019computational}
\begin{equation}
\min_{M} \langle C,M\ra\quad\text{subject to}\quad
M \geq 0,\quad
M\eins_{n} = x,\quad
M^{\T}\eins_{n} = x'.
\end{equation}
$M$ is a doubly stochastic matrix, and the minimizer $M^{\ast}$ is the optimal transport plan for transporting $x$ to $x'$, with respect to the given costs $C$.  The Wasserstein kernel is defined as 
\begin{equation}\label{eq:wasserstein_kernel}
  k_{W}(x,x')  = \exp \left( -\frac{1}{\gamma} d^{2}_{W}(x,x') \right), \quad \gamma >0,
\end{equation}
and can be shown to be a valid kernel for generating a RKHS and embedding \cite{Bachoc:2018aa}. For measures defined on the real line $\R$, it is well known that the distance $d_{W}$ between two distributions can be evaluated in terms of the corresponding cumulative distributions. This carries over to discrete measures $x, x'$ and the distance $d_{W}(x,x')$ considered here, provided the implementation takes care of monotonicity and hence invertibility of the discrete cumulative distributions; we refer to \cite[Section 2]{Santambrogio:2015aa} for details.
    
\end{description}

\subsection{Neural Networks}\label{sec:neural_networks}

\paragraph{Feed-forward neural networks.} Let $n_k$ and $n_d$ be the dimensions of the input and output space, respectively. A feed-forward neural network (FFNN) of depth $L \in \mathbb{N}$ is a function $f: \mathbb{R}^{n_k} \to \mathbb{R}^{n_d}$ that can be written as the composition $f(x)=f^{(o)}(f^{(L)}(\ldots f^{(1)}(x)))$ of $L$ hidden layers $f^{(j)}: \mathbb{R}^{n_i^{(j)}} \to \mathbb{R}^{n_o^{(j)}}, j \in [L]$, where $n_i^{(1)}=n_k$, and a final output layer $f^{(o)}:\mathbb{R}^{n_o^{(L)}} \to \mathbb{R}^{n_d}$. Each hidden layer is in turn the composition of a linear transformation and an activation function. We use  the rectified linear unit (ReLU) as activation function defined as
\begin{align}\label{eq:relu}
  \text{ReLU}(x) 
  & =
    \max (0,x), \quad x \in \mathbb{R}.
\end{align}
The hidden layers can accordingly be written as 
\begin{align}\label{eq:hidden_layer}
  f^{(j)}(x) 
  & =
    \text{ReLU}(W^{(j)}x + b^{(j)}), \quad j \in [L],
\end{align}
for an input vector $x\in \mathbb{R}^{n_i^{(j)}}$, weight matrix $W^{(j)} \in \mathbb{R}^{n_o^{(j)} \times n_i^{(j)}}$ and bias $b^{(j)} \in \mathbb{R}^{n_o^{(j)}}$. The ReLU function in Eq.~\eqref{eq:hidden_layer} acts independently on each element of its argument. For the final output function $f^{(o)}$ we use a linear transformation without activation function: 
\begin{align}
  f^{(o)}(x) 
  & =
    W^{(o)}x + b^{(o)},
\end{align}
with $W^{(o)} \in \mathbb{R}^{n_d} \times \mathbb{R}^{n_o^{(L)}}$ and bias $b^{(o)} \in \mathbb{R}^{n_d}$. The values $n_o^{(j)}$ are called the numbers of ``hidden units'' or ``neurons'' of the $j$th layer. Since for a general $W^{(j)}$ all neurons of the $(j-1)$th layer are connected to all neurons of the $j$th layer, hidden layers as in Eq.~\eqref{eq:hidden_layer} are also called ``fully-connected layers''.
To characterise a FFNN we specify the numbers of hidden units as $(n_o^{(1)},\ldots, n_o^{(L)})$. For example, $(10,20,5)$ denotes a FFNN of depth $L=3$ with $n_o^{(1)}=10, n_o^{(2)}=20$ and $n_o^{(3)}=5$, respectively. Accordingly, $()$ denotes a FFNN without any hidden layers, i.e., $L=0$.

\paragraph{Convolutional neural networks.} Convolutional neural networks (CNNs) have been used for learning problems on image data in various different applications, in particular for classification tasks \cite{gu2018recent}. We will consider CNNs that were trained on raw pattern data to learn the kinetic parameters of the model, as a benchmark for the results obtained from training models on resistance distance histograms. 

Various different CNN architectures have been used in the literature. The majority consist of three basic types of layers: convolutional, pooling, and fully connected layers. The convolutional layer's function is to learn feature representations of the inputs. This is achieved by convolving the inputs with learnable kernels, followed by applying an element-wise activation function. Convolutional layers are typically followed by pooling layers which reduce the dimension by combining the outputs of clusters of neurons into a single neuron in the next layer. Local pooling combines small clusters, typically of size $2\times 2$, while global pooling acts on all  neurons of the previous layer. A sequence of convolutional and pooling layers is then typically followed by one or several fully-connected layers as in Eq.~\eqref{eq:hidden_layer}. These are then followed by a final output layer chosen according to the specific learning task such as a softmax layer for classification tasks \cite{gu2018recent}. 

For the applications in this paper, we found the best performance for minimalistic CNNs consisting of only one convolutional and one fully-connected layer. The ReLU activation function in Eq.~\eqref{eq:relu} was applied to the output of both layers. We denote the architecture by $(n_k/ n_p/ n_f)$ where $n_k$ is the used number of kernels of size $n_p \times n_p$ and $n_f$ denotes the number of neurons in the fully connected layer.

\section{Experiments and Discussion}\label{sec:Experiments}
\subsection{Implementation Details}\label{sec:implementation_details}

\paragraph{Simulation details.}
According to Section \ref{sec:step-size-selection}, setting the step size $h$ properly requires to estimate (an upper bound of) the Lipschitz constant of $f$. It turned out, however, that applying standard calculus \cite[Ch.~9]{Rockafellar:2009aa} to the concrete mappings $f$ \eqref{eq:gm_model} yields too lose upper bounds of $L_{f}$ and hence quite small step sizes $h$, which slows down the numerical computations unnecessarily. Therefore, in practice, we set $h$ to a value that is `reasonable' for the backward Euler method and monitored the fixed point iteration \eqref{eq:fixed_point} in oder to check every few iterations if the method diverges, in which case $h$ was replaced by $h/2$. We found $h=0.2$ to be a reasonable choice for all applications studied here.

The threshold $\veps_l$ for the convergence criterion of the inner iteration in Eq.~\eqref{eq:inner_convergence_criterion} was set to $\veps_l = 0.001$. The outer iteration was terminated if either the convergence criterion in Eq.~\eqref{eq:outer_convergence_criterion} was met with threshold $\veps_{k} = 10^{-6}$, which we checked after time intervals of $\delta t =100$, or when a fixed maximal time $T_f$ was reached. We chose $T_f=2000$ for domain sizes of $32 \times 32$ and $64 \times 64$, and $T_f=5000$ for a domain size of $128 \times 128$. We found that the patterns do not change substantially beyond these time values even if the convergence criterion in Eq.~\eqref{eq:outer_convergence_criterion} was not satisfied. 

\paragraph{Resistance distance histograms.} As pointed out in Remark \ref{rem:redundancy_different_species}, the resistance distance histograms (RDHs) for different species are typically redundant. We hence used only the first species' simulation results for computing the RDHs from simulations of the Gierer-Meinhardt model in Eq.~\eqref{eq:gm_model} studied here.

For computing RDHs, we had to specify the edge weight parameter $\epsilon$ of \eqref{eq:edge_weights} which penalises paths from high to low concentrations and vice versa. Choosing $\epsilon$ too small (corresponding to a large penalty) leads to a saturation effect of large resistance values between nodes at large distances, preventing to resolve the geometry of a pattern on such larger scales. Similarly, a large $\epsilon$ fails to resolve the geometry on small scales. We empirically found $\epsilon = 0.003$ to be a good compromise. 

For the parameter $t$ in Definition \ref{def:RDH} determining the undersampling of the graph, we found $t=1$ to give the most accurate results. Thus, all results presented in this study were produced using $t=1$. Note that $t=1$ means that the original graph was used without undersampling. 

When simulating a model for varying parameters, we found that some patterns led to few occurences of very large resistance values, while the majority of patterns had maximal resistance values substantially below these outliers. We believe that these large values arise from numerical inaccuracies in the PDE solver. Rather than including all resistance values which would cause most RDHs having only zeros for large values, we disregarded values beyond a certain threshold. To specify this threshold, we computed the $99\%$ quantile across all patterns and picked the maximal value. 

Finally, we set the bin number $B$ introduced in Definition \ref{def:RDH} to the value $B=12$. We found empirically that smaller bin numbers give more accurate results for small data sets, while larger numbers perform better for larger data sets. $B=12$ appears as a good tradeoff between these two regimes.

\paragraph{Additional features.}
 
In Section \ref{sec:additional_features} we discussed the maximal concentration as an additional feature. Due to numerical inaccuracies when simulating a system, a few pixels might have an artificially large concentration. We aim here to disregard such values and instead estimate the concentration value of the highest plateau in a given pattern. To this end, we collected the concentration values of the pattern into a histogram of 25 bins and defined the maximal concentration as the location of the peak with the highest concentration value.

\paragraph{Data splitting.}

Consider the learning problem described in Section \ref{sec:sub_learning_problem}: given a data set $\mc{D}_{m} = \{(x_i, y_i)\}_{i\in [m]} \subset \mc{X}\times\mc{Y}$ with RDHs $x_i$ and vectors $y_i$ comprising parameter values of the model, we aimed to learn a prediction function $f:\mc{X} \to \mc{Y}$. In practice, we did not use the whole data set $\mc{D}_{m}$ for training, but split it into mutually disjoint training, test and validation sets, which respectively comprised $60\%, 20\%$ and $20\%$ of $\mc{D}_{m}$. The training set was used to train a model for given hyperparameters, while the validation set was in turn used to optimise the hyperparameters. The NRMSE of the trained model was subsequently computed on the test set.
 
For small data sets $\mc{D}_{m}$ with $m\leq 500$, we observed large variations in the resulting NRMSE values. To obtain more robust estimates we split a total data set of $1000$ points into subsets $\mc{D}_{m}$ of size $m$ for $m \leq 500$, performed training and computed the NRMSE value for each $\mc{D}_{m}$ as described above, and took the average over these NRMSE values. For example, if $m=100$, then we averaged over $1000/100=10$ data sets. For data sets of size $m>500$, the procedure above was performed on the single data set without averaging.

For convolutional neural networks trained on raw pattern data we found the results to be substantially more noisy than for the other learning methods trained on RDHs. Accordingly, we here averaged the NRMSE values over more training sets: for $m\leq 1000$ points, we split a total set of $5000$ points into data sets $\mc{D}_{m}$. For $m=2000, 5000$ and $10000$ we used total data sets of $20000$ points. Finally, for $m=20000$ we trained the models three times with random initialization on the same data set and averaged subsequently.

\paragraph{Target variable preprocessing.}

We normalised each component of the target variable by its maximal value over the whole data set, i.e., 
\begin{align}
  y_{i,j}'
  & =
    \frac{y_{i,j}}{\max \{y_{l,j}\}_{l \in [n]}}, \quad i \in [n], \quad j \in [d],
\end{align}
where $d$ is the dimension of the target variable corresponding to the number of parameters to be learned, and $n$ is the number of data points. We use these normalised target variables for both the regression task in Section \ref{sec:learning_parameters} and for clustering in Section \ref{sec:clustering}

\paragraph{Support-Vector Regression.}

The training procedure for learning a single parameter, i.e. a scalar-valued target variable, using support-vector regression (SVR) is described in Section \ref{sec:svr-separable}. We choose the hyperparameters $\gamma$ (c.f.~Eq.~\eqref{eq:chi_squared_kernel_exp} for the exponential $\chi^2$-kernel and Eq.~\eqref{eq:wasserstein_kernel} for the Wasserstein kernel) and $\lambda$ (c.f.~Eq.~\eqref{eq:svr_training_problem}) by minimising the NRMSE on the validation set on a grid in the two parameters (note that the $\chi^2$-kernel of Eq.~\eqref{eq:chi_squared_kernel} does not contain any hyperparameter). In some cases, we performed a second optimisation over a finer grid centered around the optimal parameters from the first run. We found this to lead to only minor improvements, however. 
The model was then evaluated on the test set for the optimal parameters and the resulting NRMSE value is reported. 

For learning multiple parameters we applied the SVR approach separately to each target parameter and subsequently computed the joint NRMSE value.

\paragraph{Operator-Valued Kernels.}

For learning multiple parameters jointly, i.e. a vector-valued target variable, we used the operator-valued kernel method described in Section \ref{sec:operator-valued_kernel}. In addition to the input kernel parameter $\gamma$ and regression parameter $\lambda$ used for support-vector regression, we here also had to optimise the scale parameter of the output kernel (c.f.~the discussion after Eq.~\eqref{eq:preimage-problem-Ln}). Optimisation of these hyperparameters was performed on a grid as in the SVR case, but this time jointly for all target parameters.

\paragraph{Feed-forward neural networks.} We employed feed-forward neural networks both for learning a single parameter as well as learning multiple parameters jointly from RDHs.
We used $\mathrm{Mathematica's}^{\copyright}$ build-in \texttt{NetTrain} function with the Adam optimization algorithm and the mean-squared loss function for training \cite{Mathematica}. The network architecture that gives the minimal loss on the validation set along the training trajectory was selected and evaluated on the test set to obtain the NRMSE value. 
Training was performed for $T_f$ training steps with early stopping if the error on the validation set does not improve for more than $T_e$ steps. For the data set sizes $20$ and $50$ we used $(T_f,T_e)$ = $(4\times 10^5, 10^5)$, for data set sizes $100,1000$ and $2000$ we used     $(T_f,T_e)$ = $(2\times 10^5, 5 \times 10^4)$, and for data set sizes
 $\geq 5000$ we used $(T_f,T_e)$ = $(10^5, 2 \times 10^4)$.

\paragraph{Convolutional neural networks.} Convolutional neural networks were trained on the raw simulation data of the first species, i.e. \emph{not} on RDHs. We used TensorFlow \cite{tensorflow2015-whitepaper} and Keras \cite{chollet2015keras} for this purpose. We employed the same training procedure as for feed-forward neural networks. For the number of training steps we used $(T_f,T_e)=(500,100)$.

\begin{figure}[t]
\begin{center}
\centerline{
\includegraphics[width=0.8\textwidth]{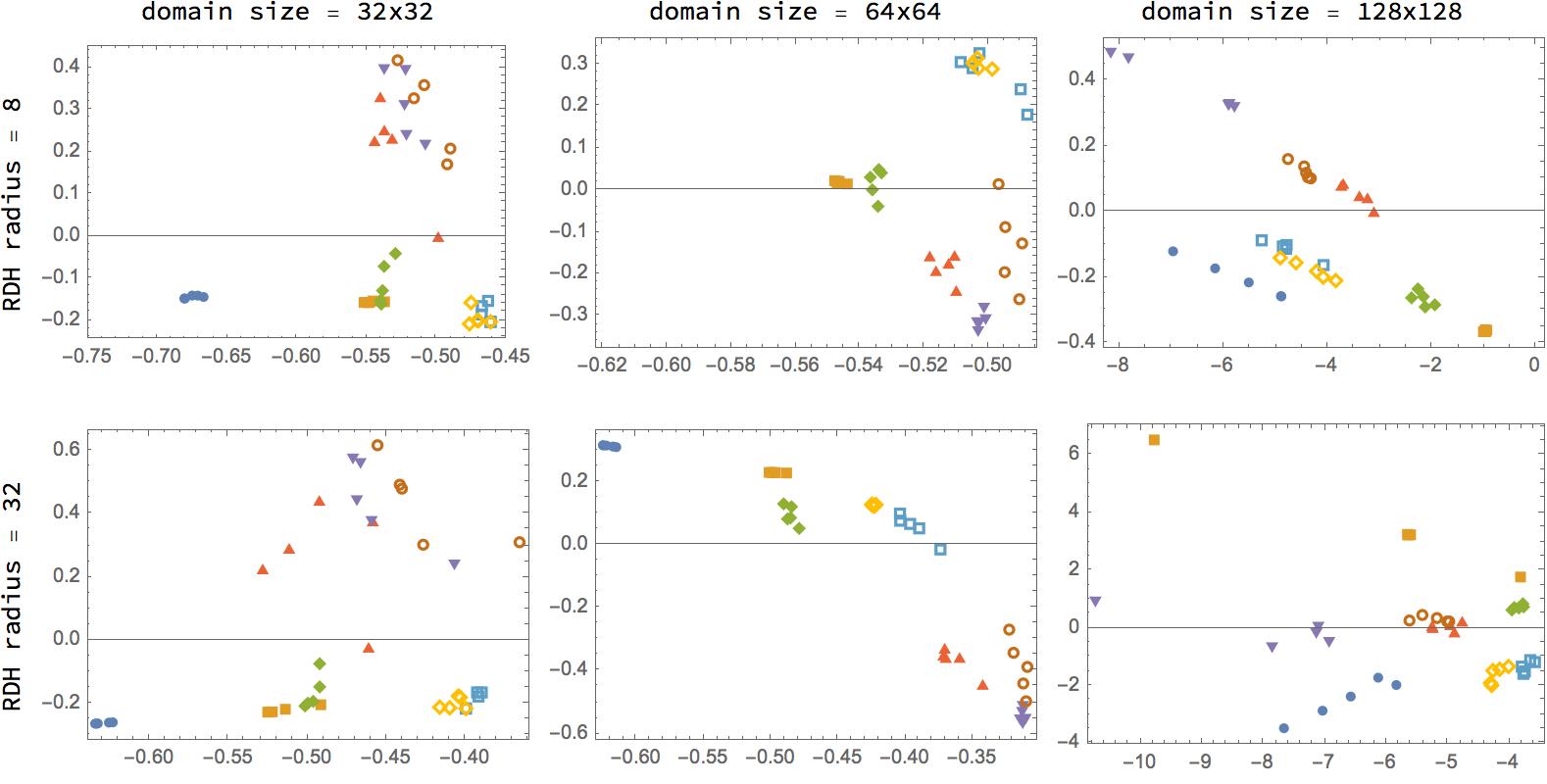} 
\hspace{0.2cm}
\includegraphics[width=0.2\textwidth]{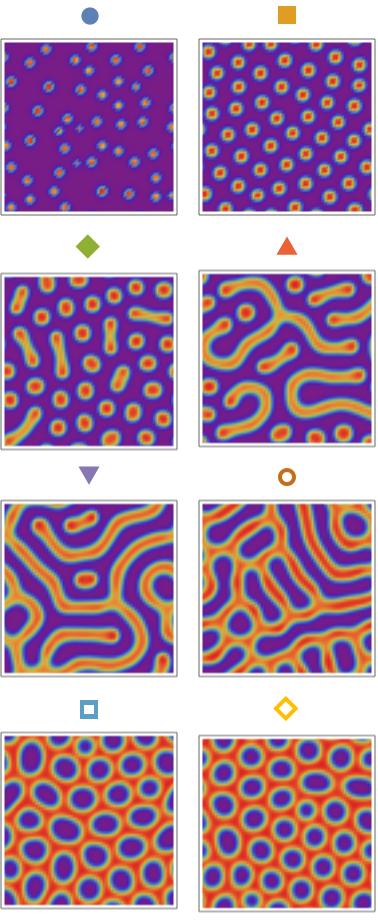}
}
\centerline{
\parbox{0.8\textwidth}{\centering (a)} \hfill
\parbox{0.2\textwidth}{\centering (b)}
}
\caption{(a) Dimensionality reduction of simulation results of the Gierer-Meinhardt model defined in Eq.~\eqref{eq:gm_model} for eight equally-distanced values of parameter $c$ on the interval $[0.01, 1.15]$
indicated by different symbols, and five different random initial conditions each. The other parameters are fixed to $a=0.02, b=1, \delta=100$.
The resulting RDHs are embedded into the two-dimensional plane using latent semantic analysis \cite{Berry:1995wp}. Results are shown for domain sizes $32 \times 32$, $64 \times 64$ and $128 \times 128$ and radii $8$ and $32$. We found that while points corresponding to different patterns do not separate into distinct clusters for a domain size of $32 \times 32$, they do so for a domain size of $64 \times 64$. For a domain size of $128 \times 128$, this separation appears even more pronounced. This demonstrates that resistance distance histograms successfully encode characteristic features of patterns while averaging out noise, if the domain size is chosen large enough.
(b) Patterns for the eight different $c$ values shown in (a) for one initial condition each. 
 }\label{fig:dimensionality_reduction_of_rdhs}
\end{center}
\end{figure}
%

\subsection{Robustness of Resistance Distance Histograms}\label{sec:dim_reduction}

Ideally, the RDHs should be characteristic of patterns of different types, while being robust to noise in the patterns due to noise in the initial conditions, and being invariant under immaterial spatial pattern transformations (translation, rotation). In other words, patterns arising from simulations of the same model for different initial conditions should give rise to RDHs that do not differ substantially, while patterns generated by different models should lead to larger deviations. 

To assess these properties, we simulated the Gierer-Meinhardt model introduced in Section \ref{sec:examples} for eight different values for $c$ with the other parameters fixed. For each value of $c$ we simulate the model for five random initial conditions. We subsequently embed the corresponding RDHs into two dimensions.
Figure \ref{fig:dimensionality_reduction_of_rdhs} shows the results for differing domain sizes. We observe that the points are reasonable well clustered for a domain size of $32\times 32$ pixels, with a substantial improvement when increasing the domain size to $64 \times 64$ pixels. This is to be expected, since a larger snapshot of a pattern should reduce the noise in the corresponding RDHs. The clustering appears to improve slightly upon further increase of domain size to $128 \times 128$ pixels. 

The fact that the different noisy realisations of the patterns separate well indicates that the RDHs average out this noise while encoding the characteristic features of the patterns to a large degree. 

In view of these results, we only consider domain sizes of $64 \times 64$ and $128 \times 128$ in the following.

\begin{figure*}[t]
\begin{center}
\centerline{\includegraphics[width=1\textwidth]{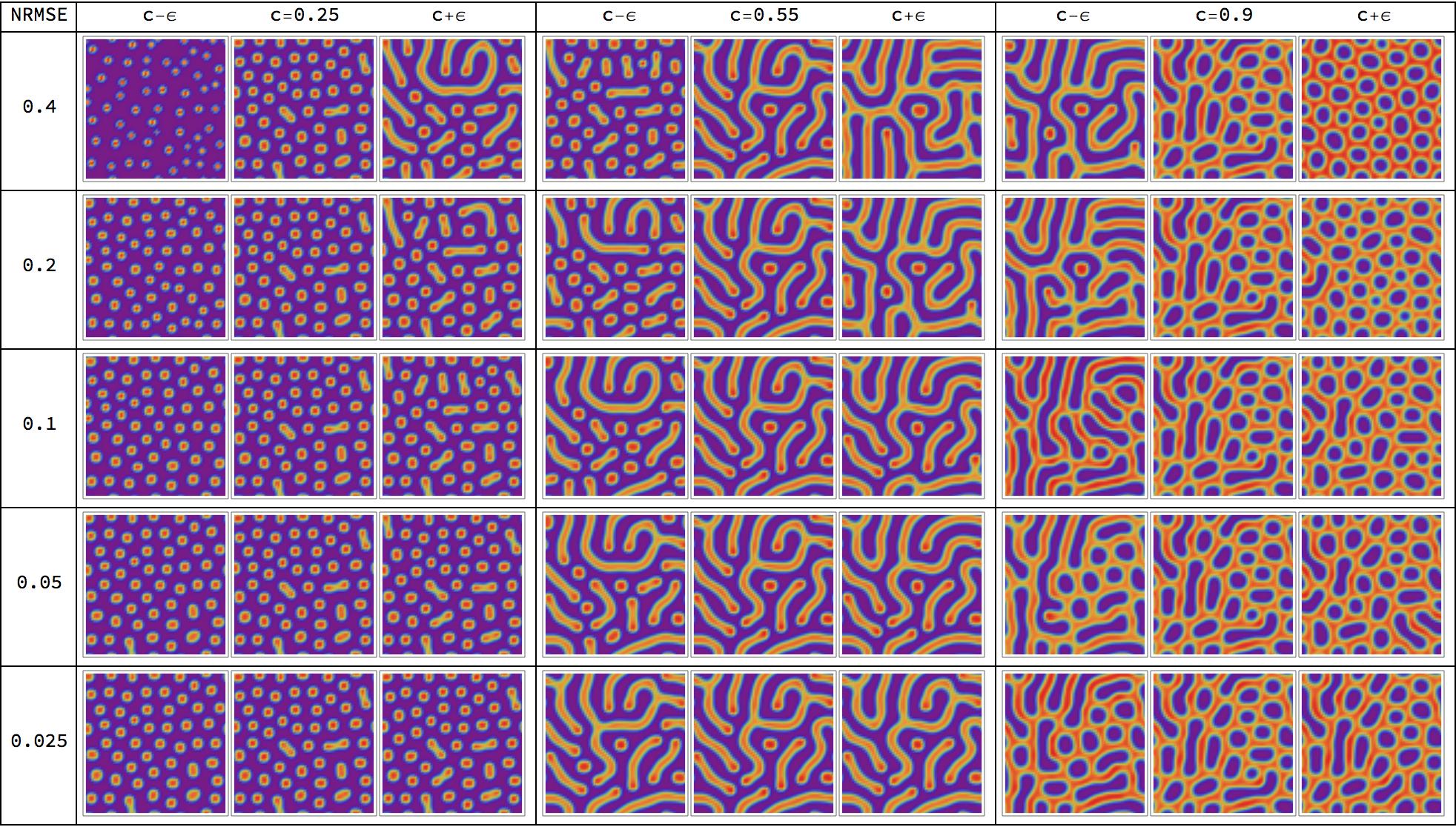}}
\caption{\textbf{Pattern accuracy.} Simulation results of the Gierer-Meinhardt model in Eq.~\eqref{eq:gm_model} for fixed parameters $a=0.02, b=1$ and $\delta=100$ and varying values for parameter $c$ on a $64 \times 64$ domain. The $c$ values are varied around a central value such that they correspond to a certain NRMSE value, and different rows correspond to different NRMSE values. We find that for an NRMSE value of $0.4$ the patterns deviate quite substantially from each other, while they look relatively similar for a value of $0.2$ already. Decreasing the NRMSE value further successively decreases the deviations in the patterns. For NRMSE values smaller than $0.05$ different patterns are hardly distinguishable anymore.}\label{fig:patterns_for_varying_nrmse_values}
\end{center}
\end{figure*}
%

\subsection{Learning Parameters of the Gierer-Meinhardt Model}\label{sec:learning_parameters}

In this section we consider the prediction problem of learning a map from RDHs (potentially combined with the additional features described in Section \ref{sec:additional_features}) generated from simulations of the Gierer-Meinhardt model in Eq.~\eqref{eq:gm_model} as described in Section \ref{sec:numerical_simulation}  onto the corresponding kinetic parameters. The training data thus consists of pairs $(x_i, y_i)$ with $x_i$ being an RDH and $y_i$ being a set of kinetic parameters of the model. As outlined in Section \ref{sec:implementation_details} we split the data into a training, validation and test set, where the former two are used to train the models and learn hyperparameters, and the latter is used to evaluate the model's error in terms of the normalised root-mean squared error (NRMSE) (c.f.~Section \ref{sec:accuracy_measure}). 

Figure \ref{fig:patterns_for_varying_nrmse_values} visualises how patterns vary for varying parameters corresponding to different NRMSE values. We observe that the patterns look reasonably similar for NRMSE values of $0.2$, while they are hardly distinguishable anymore for values below $0.05$.

\begin{figure*}[t]
\begin{center}
\centerline{
\includegraphics[width=0.48\textwidth]{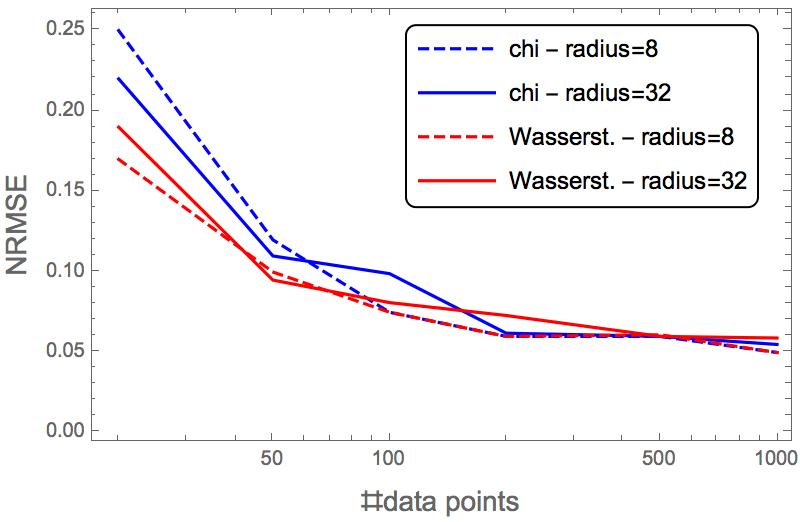}
\hspace{0.3cm}
\includegraphics[width=0.48\textwidth]{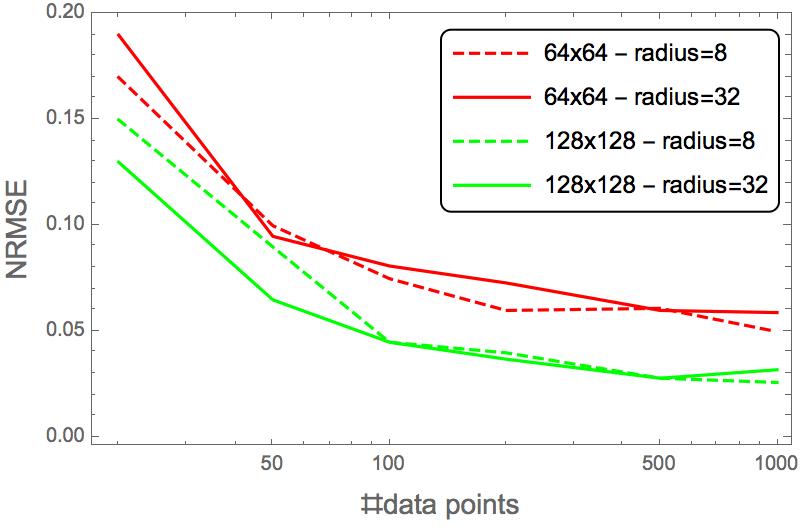}
}
\centerline{
\parbox{0.48\textwidth}{\centering (a)}
\hspace{0.3cm}
\parbox{0.48\textwidth}{\centering (b)}
}
\caption{NRMSE values for support-vector regression of parameter $c$ of the Gierer-Meinhardt model in Eq.~\eqref{eq:gm_model}. We vary $c$ uniformly on the interval $[0, 1.15]$ and fix the other kinetic parameters to $a=0.02, b=1$ and $\delta=100$, the scaling parameter to $s=0.25$. The figures show the NRMSE values for varying data set sizes and for the two RDH-radii $8$ and $32$ (c.f.~Definition \ref{def:RDH}). (a) The figure shows the results for both the exponential $\chi^2$- and Wasserstein kernel (c.f.~Eqs.~\eqref{eq:chi_squared_kernel} and \eqref{eq:wasserstein_kernel}, respectively) and a domain size of $64 \times 64$. We find that even for small data sets of only $20$ data points a reasonable good NRMSE value of about $0.2$ is achieved (c.f.~Figure \ref{fig:patterns_for_varying_nrmse_values}). This value successively decreases for increasing data set sizes down to a value of about $0.05$. While the results for the two different RDH-radii do not vary substantially, the Wasserstein kernel outperforms the $\chi^2$ kernel for small data sets.
(b) The figure shows the results for the Wasserstein kernel in Eq.~\eqref{eq:wasserstein_kernel} and the two domain sizes $64 \times 64$ and $128 \times 128$. We observe about about $10-50\%$ better results for the $128 \times 128$ domain. These results show that about $1000$ data points suffice to reach the NRMSE value $0.05$ which is quite accurate (cf.~Figure \ref{fig:patterns_for_varying_nrmse_values}).
}
\label{fig:svr_scalar}
\end{center}
\end{figure*}
%

\subsubsection{Learning a Single Parameter}\label{sec:learning_single_parameter}

We start by varying the parameter $c$ and fixing the other parameters to $a=0.02, b=1$ and $\delta=100$ (c.f.~Eq.~\eqref{eq:gm_model}). We randomly sample $2 \times 10^4$ values for $c$  on the interval $[0, 1.15]$, solve the corresponding PDE in Eq.~\eqref{eq:gm_model} and compute the resulting RDHs as described in Section \ref{sec:resistance_histograms}. Several different types of patterns emerge in this range of $c$ values as can be seen in Figure \ref{fig:dimensionality_reduction_of_rdhs}(b).

\paragraph{Support-vector regression.}

Figure \ref{fig:svr_scalar}(a) shows the NRMSE obtained by training the support-vector regression model with both the exponential $\chi^2$-kernel and the Wasserstein kernel as introduced in Section \ref{sec:kernel-functions}, for the two RDH-radii $8$ and $32$. We find that small data sets of only $20$ data points allow to learn the parameter $c$ reasonable well with NRMSE values in the range $0.17-0.25$, which indicates that the RDHs average out noise in the patterns to a large degree (as already noted in Section \ref{sec:dim_reduction}).  

Increasing the number of data points successively reduces the NRMSE down to values of $0.059-0.055$ for $1000$ data points. We observe that even for relatively small snapshots of the patterns of only $64 \times 64$ pixels, the RDHs allow to learn the parameter $c$ with quite high accuracy. While we don't observe a substantial difference between the two analysed RDH-radii we do find that the Wasserstein kernel gives more accurate results for small data sets, while the two kernels perform similar for larger data sets. This finding is plausible because unavoidable binning effects like slightly shifted histogram entries impact RDHs more when the data set is small, but are reasonably compensated through `mass transport' by the Wasserstein kernel. We ran the same experiments for the symmetric $\chi^2$ kernel introduced in Section \ref{sec:kernel-functions} and obtained worse results than for the other two kernels (results not shown).  
Therefore, in the rest of this paper, we will use the Wasserstein kernel. 

Figure \ref{fig:svr_scalar}(b) shows the results for an increased domain size of $128 \times 128$. This larger domain leads to improved NRMSE values of about $10-50\%$, with a larger improvement for larger data sets. As already noted in Section \ref{sec:dim_reduction}, this improvement is to be expected since a larger snapshot of a pattern allows the RDHs to average out local fluctuations more efficiently. 

For simplicity and computational convenience, however, we use only $64 \times 64$ domain sizes in the following.

\begin{figure}

\begin{minipage}{0.45\textwidth}
\begin{figure}[H]
    \centering
    \includegraphics[scale=0.45]{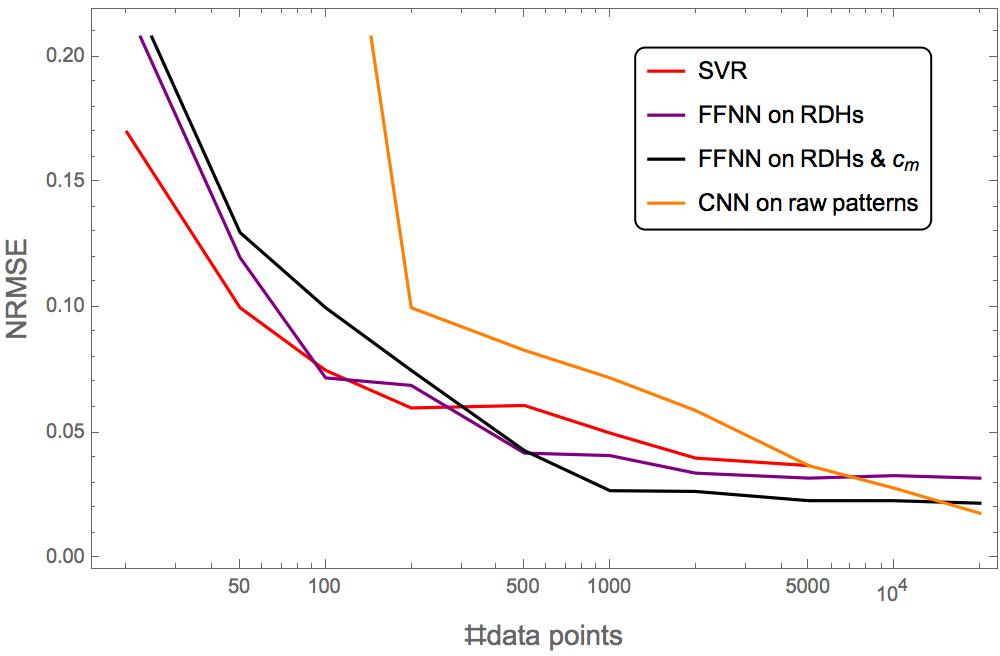}
\end{figure}

\end{minipage}
\hfill
\begin{minipage}{0.5\textwidth}
\begin{table}[H]
    \centering
    \tabcolsep=0.11cm
    \footnotesize
    \begin{tabular}{|r|c|c|c|}
    \hline
        \begin{tabular}{@{}c@{}}\#data  \\ points\end{tabular} &  
            \begin{tabular}{@{}c@{}}FFNN on  \\ RDHs\end{tabular}  &  
            \begin{tabular}{@{}c@{}}FFNN on  \\ RDHs \& $c_m$\end{tabular}   & 
            \begin{tabular}{@{}c@{}}CNN on  \\ raw patterns \end{tabular} \\
        \hline
        20 & () & () & - \\
        50 & () & () & - \\
        100 &  (2) & () & (5/5/5) \\
        200 &  (2) & (2) & (5/5/5) \\
        500 &  (20) & (20) & (5/5/5) \\
        1000 &  (5,10) & (10,10) & (5/5/5) \\
        2000 &  (10,10) & (20) & (5/5/5) \\
        5000 &  (10,10) & (20,20) & (10/5/5) \\
        10000 &  (20,20) & (20,20) & (10/5/5) \\
        20000 &  (20,20) & (20,20) & (5/5/15) \\
        \hline

    \end{tabular}

\end{table}
\end{minipage}

\centerline{
\parbox{0.55\textwidth}{\centering (a)}
\parbox{0.35\textwidth}{\centering (b)}
}

\caption{(a) NRMSE values for the same setting as in Figure \ref{fig:svr_scalar}, for a RDH radius of $8$ and a domain size of $64 \times 64$. The figure shows the results obtained using support-vector regression (SVR) trained on RDHs, feed-forward neural networks (FFNNs) once trained on RDHs and once trained on RDHs combined with the maximal concentration $c_m$ as additional feature (c.f.~Section \ref{sec:additional_features}), as well as convolutional neural networks (CNNs) trained on the raw patterns, as described in in Section \ref{sec:neural_networks}. We observe that the FFNNs perform slightly worse than the support-vector regression for small data sets, similar for intermediate data set sizes of $100-200$ points, and slightly better for larger data sets. The FFNNs trained on RDHs and the maximal concentration $c_m$ perform slightly worse for less than $500$ data points and slightly better for larger sets.
The NRSMSE value seems to level off and not decrease any further for larger data sets. 
For the CNNs trained on the raw patterns we find that NRMSE values are substantially larger than the corresponding FFNN and SVR values, which one may expect due to overfitting. The NRMSE values lie outside of the shown plot range for data sets smaller than $200$ points. The difference decreases for increasing data sets until the CNNs eventually become more accurate for $10^4 - 2 \times 10^4$ data points. 
Note that no SVR results for $\geq 10^4$ data points are shown since our basic QP-solver failed to converge. However, the SVR method only outperforms the other methods for quite small data sets 
anyway.
(b) Architectures for both the FFNNs and CNNs that gave the best performance and whose results are shown in (a) (see Section \ref{sec:neural_networks} for the used notation).
}
\label{fig:svr_vs_nn_scalar}
\end{figure}

\paragraph{Neural networks.}
Figure \ref{fig:svr_vs_nn_scalar} shows the regression results for learning the parameter $c$ using feed-forward neural networks (FFNN) for data sizes of up to $2\times 10^4$ points. As one may expect, the FFNNs perform worse than support-vector regression for small data sets and better for larger data sets. Using FFNNs it is feasible to use data sets beyond the maximum of $5000$ points used for support-vector regression. However, we find that the NRSME value appears to not improve any further beyond about $2000$ points. This may be expected since in the computation of the RDHs  some information about the patterns is inevitably lost, meaning there is a lower bound of how accurate the parameter can be learned in the limit of an infinitely large data set. We point out, however, that this saturation effect happens at NRSME values $\leq 0.03$ which is very accurate (cf.~Figure \ref{fig:patterns_for_varying_nrmse_values}).

\begin{figure}
\begin{minipage}{0.6\textwidth}
\begin{figure}[H]
    \centering
    \includegraphics[scale=0.5]{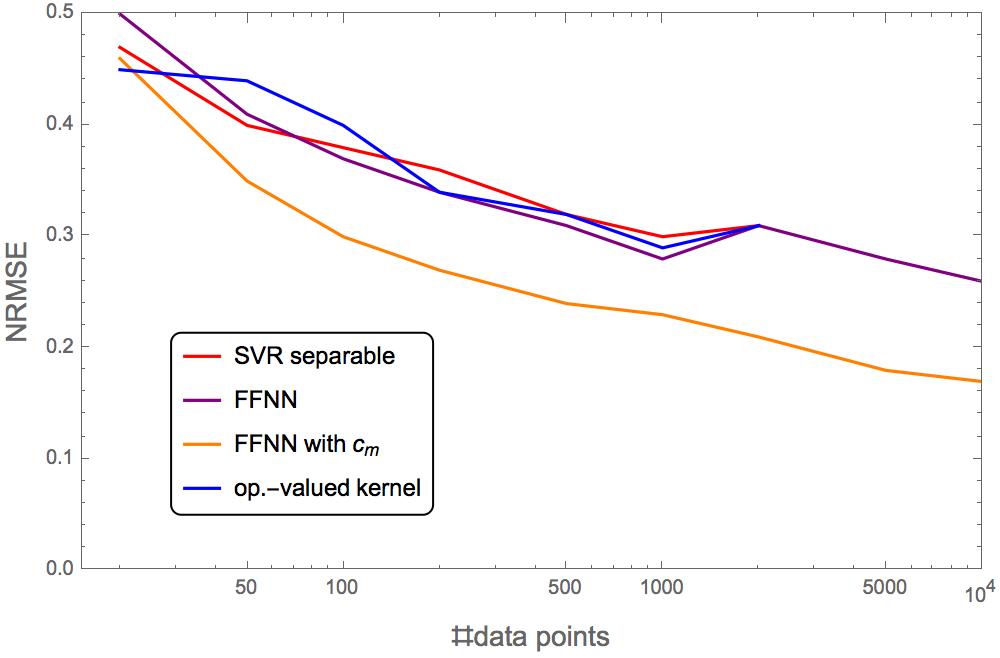}
\end{figure}
\end{minipage}
\hfill
\begin{minipage}{0.4\textwidth}
\begin{table}[H]
    \centering
    \tabcolsep=0.11cm
    \footnotesize
    \begin{tabular}{|r|c|}
    \hline
        \begin{tabular}{@{}c@{}}\#data  \\ points\end{tabular} &  
            \begin{tabular}{@{}c@{}}FFNN  \\ architecture \end{tabular} \\
        \hline
        20 & () \\
        50 & ()   \\
        100 & ()   \\
        200 & ()   \\
        500 & (5,5)   \\
        1000 & (5,5)   \\
        2000 & (10,10)   \\
        5000 & (20,20)   \\
        10000 & (20,20)   \\
        \hline
    \end{tabular}
\end{table}
\end{minipage}
\centerline{
\parbox{0.6\textwidth}{\centering (a)}
\parbox{0.35\textwidth}{\centering (b)}
}
\caption{(a) NRMSE values for learning all four kinetic parameters $a,b,c$ and $\delta$ of the Gierer-Meinhardt model in Eq.~\eqref{eq:gm_model}.
The figures show the NRMSE values for varying data set sizes and for the RDH-radius $8$ (c.f.~Definition \ref{def:RDH}) and a domain size of $64 \times 64$, for both cases of kernel-based learning the four parameters individually and jointly as outlined in Sections \ref{sec:svr-separable} and \ref{sec:operator-valued_kernel}, respectively, using the Wasserstein kernel (c.f.~Eqs.~\eqref{eq:wasserstein_kernel}). In addition, the NRMSE values of feed-forward neural networks (FFNNs) (c.f.~Section \ref{sec:neural_networks}) are shown, once trained only on RDHs and once trained using the maximal concentration $c_m$ as additional feature (c.f.~Section \ref{sec:additional_features}).  As one may expect, the NRMSE values are substantially larger here than in the scalar case of learning just one parameter for the same number of data points (c.f.~Figure \ref{fig:svr_vs_nn_scalar}) and once again decrease for increasing data sets. 
We find that all three methods trained on RDHs perform very similar. This implies, in particular, that no correlation among the output parameter values could be exploited for prediction. In contrast, including the maximal concentration $c_m$ as an additional feature leads to substantially improved NRMSE values with an improvement of up to $35\%$ for large data sets. 
(b) Architectures for the FFNNs that gave the best performance and whose results are shown in (a). We found the same optimal architectures for both training the FFNNs on the RDHs only and on the RDHs together with the maximal concentration $c_m$.
}
\label{fig:svr_vs_nn_vector}
\end{figure}

\paragraph{Additional features.}
In Section \ref{sec:additional_features} we introduced two additional features to account for certain symmetries of the RDHs, namely the maximal concentration $c_m$ and the number of connected components $n_c$ of a pattern. Figure \ref{fig:svr_vs_nn_scalar} shows the results obtained by training FFNNs on RDHs with $c_m$ as an additional feature. 
We find slightly larger NRMSE values for small data sets with less than $500$ points, and slightly smaller NRMSE values for larger data sets. In contrast, using the number of connected components $n_c$ as an additional feature did not give rise to notably more accurate results (results not shown).

\paragraph{Benchmark: CNN on raw data.} Figure \ref{fig:svr_vs_nn_scalar} also shows the NRMSE values obtained from training convolutional neural networks (CNNs) \textit{directly on the raw pattern data} obtained through simulation (c.f.~Section \ref{sec:neural_networks}). For data set sizes of $\leq 200$ data points we found substantially larger NRMSE values than from FFNNs trained on RDHs. This shows once again that the RDHs efficiently encode most of the relevant information while averaging out noise, allowing for more accurate parameter learning for data sets of small and medium size. As one might except, the difference between the CNN and FFNN results becomes smaller for larger data set sizes since the CNNs can effectively average out the noise themselves when a sufficient large number of data points are provided. Consequently, we found that CNNs become more accurate than the other methods for large data sets of about $10^4 - 2 \times 10^4$ data points.

\subsubsection{Jointly Predicting Four Parameters}

We next consider the problem of learning all four kinetic parameters $a,b,c$ and $\delta$ of the Gierer-Meinhardt model in Eq.~\eqref{eq:gm_model}. We vary $a,b,c$ and $\delta$ uniformly on the interals $[0.01, 0.7]$, $[0.4,2]$, $[0.02,7]$ and $[20,200]$, respectively. The scaling parameter was set to $s=0.4$ and assumed to be known. Parameter combinations for which the system does not possess a Turing instability and hence does not produce a pattern in simulations are disregarded. 

Figure \ref{fig:svr_vs_nn_vector} shows the NRMSE results for the separable SVR model, for the joint kernel-based model, and for feed-forward neural networks (FFNNs) trained on RDHs of radius $r=8$ and a domain size of $64 \times 64$. We find that the three methods perform similar which means, in particular, that no correlation among the output parameters could be exploited for joint prediction in order to outperform separable parameter prediction (the SVR methods we trained only for data sets of up to $2 \times 10^3$ points).
We further observe the NRMSE values to be substantially higher than in the scalar case (c.f.~Figure \ref{fig:svr_vs_nn_scalar}) for the same number of data points, which is to be expected when learning more parameters. The NRMSE value again successively decreases for increasing data set sizes, with a minimal value of about $0.26$ for FFNNs and $10^4$ data points, which is substantially larger than the minimal value of $0.033$ obtained in the scalar case for the same data set size and radius (c.f.~Figure \ref{fig:svr_vs_nn_scalar}). However, here the curve does not appear to have levelled off yet for $10^4$ data points as in the scalar case, and increasing the data set size further should further increase accuracy. 

 We performed the same experiment as shown in Figure \ref{fig:svr_vs_nn_vector} but for RDHs radius $r=32$ and found similar results (results not shown).  

\paragraph{Combining RDHs of different radii.}
In Section \ref{sec:Feature-extraction} we argued that the RDH radius $r$ determines the scale at which the local structure of a Turing pattern is resolved. We used the two radii $r=8$ and $r=32$ in the results presented so far in Figures $\ref{fig:svr_scalar}-\ref{fig:svr_vs_nn_vector}$ and found similar results for the two. However, since RDHs with radius $r=8$ should more accurately capture characteristics of patterns on small scales and $r=32$ should be able to capture larger-scale characteristics, one might expect that combining the two should provide more information than each of them individually and might therefore give rise to more accurate results. We trained feed-forward neural networks on the RDHs of the two radii taken together as features for the same setting as in Figure \ref{fig:svr_vs_nn_vector}, but did not obtain notably more accurate results (results not shown).  

\paragraph{Additional features.}

While we found in Section \ref{sec:learning_single_parameter} that using the maximal concentration $c_m$ as an additional feature did only slightly improve NRMSE values and only for large data sets when learning a single parameter (c.f.~Figure \ref{fig:svr_vs_nn_scalar}), we here find a substantial improvement for all data set sizes, with improvements of up to $35\%$ for large data set sizes as can be seen in Figure \ref{fig:svr_vs_nn_vector}.
As in the scalar case, we find that using the number of connected components $n_c$ does not improve results (results not shown).

\begin{figure*}[t]
\begin{center}
\centerline{\includegraphics[width=\textwidth]{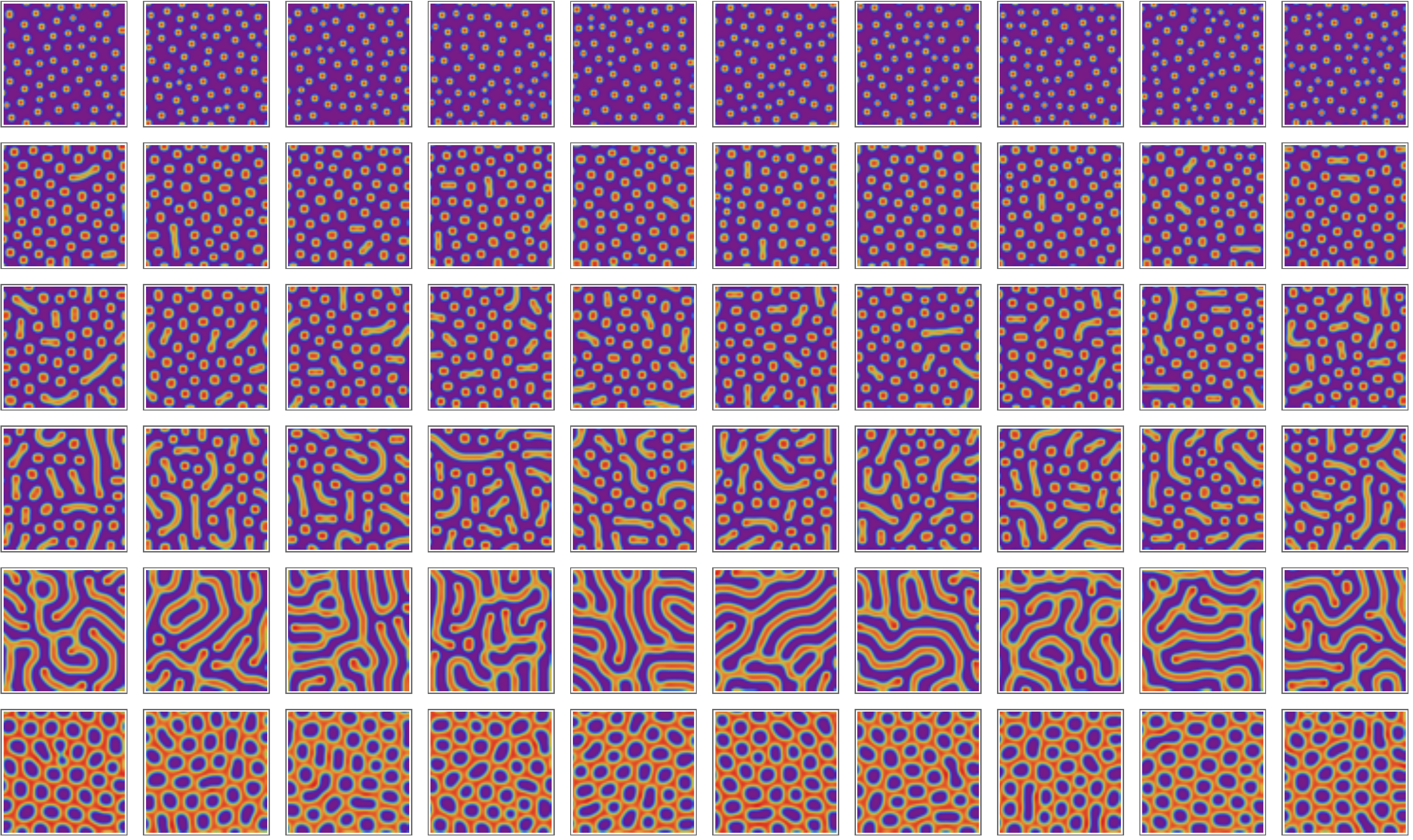}}
\caption{Cluster of patterns obtained by varying parameter $c$ as described in Section \ref{sec:learning_single_parameter}. Each row shows 10 sample patterns from a cluster that comprises a large number of patters within a small radius, measured by the squared Wasserstein distance \eqref{eq:dW2} between the corresponding resistance distance histograms (RDHs). This result demonstrates how RDHs and a corresponding distance function represent distinct types of patterns. 
}
\label{fig:clustering}
\end{center}
\end{figure*}
%

\subsection{Cluster of Patterns}\label{sec:clustering}

We explored the geometry of patterns represented by resistance distance histograms (RDHs) and the squared Wasserstein distance \eqref{eq:dW2}. Parameter $c$ in the Gierer-Meinhardt model in Eq.~\eqref{eq:gm_model} was varied as described in Section \ref{sec:learning_single_parameter}, and 1000 patterns and corresponding RDHs were computed as detailed in Section \ref{sec:numerical_simulation}. Next, we examined the neighborhood graph in which two 
patterns with corresponding RDHs $x, x'$ are adjacent if $d_{W}^{2}(x,x')\leq 0.05$. As a result, about $84\%$ of all patterns were contained in one of the six clusters corresponding to the connected components with the largest number of patterns.

Figure \ref{fig:clustering} depicts 10 sample patterns taken from each cluster. This result shows that RDHs according to Definition \ref{def:RDH}, together with an appropriate distance function, are suited for clustering patterns into qualitatively different categories.

\section{Conclusion}\label{sec:Conclusion}
We introduced a novel learning-based approach to Turing pattern parameter prediction. A key difference to existing work is that any \textit{single} observed pattern is \textit{directly} mapped to a predicted model parameter value. Major ingredients of the method are (i) the (almost) invariant representation of Turing patterns by histograms of resistance distances computed within patterns, and (ii) a kernel-based pattern similarity measure based on the Wasserstein distance that takes properly into account minor but unavoidable binning effects. 

We compared classical reproducing kernel Hilbert space methods using basic kernels for single parameter prediction and operator-valued kernels for jointly predicting all parameters. These methods performed best for small and medium-sized training data sets. In addition, we evaluated various feed-forward neural network architectures for prediction. As for single parameter prediction, these methods performed best for larger training data sets and but were on a par only with kernel-based methods in the case of joint parameter prediction.

Finally, we applied convolutional neural networks to raw pattern data directly. We found that, for very large training data sets with $\geq 2\cdot 10^4$ data samples, they outperformed all other methods. However, it remains unexplained what internal pattern representations are used.

Overall, we observed excellent parameter prediction of single parameters even for small data sets with $\leq 1.000$ data samples, and fairly accurate joint prediction of all parameters for large data sets. Our results indicate that the latter predictions should further improve when even larger data sets can be used for training. We leave such experiments for future work.

\vspace{0.25cm}
We suggest to focus on two aspects in future work. In this paper, the Gierer-Meinhardt model was chosen in order to conduct a representative case study. In practical applications, selecting also the model among various candidates, besides estimating its parameters, might be desirable. Furthermore, our current approach cannot quantify the uncertainty of parameter prediction and in this respect falls short of statistical approaches like \cite{campillo2019bayesian,kazarnikov2020statistical}. On the other hand, our approach can be applied to single patterns, rather than to ensemble of patterns like the approach \cite{kazarnikov2020statistical}, and further input data like initial conditions as in \cite{campillo2019bayesian} that are unknown in practice, are not required. Resolving these pros and cons defines an attractive program for future research.

\vspace{0.5cm}
\noindent
\textbf{Acknowledgements.} DS gratefully acknowledges support from a “Life?” programme grant from the Volkswagen Stiftung and the BBSRC through Grant BB/P028306/1.

\bibliographystyle{amsalpha}
\bibliography{learning_turing}

\newcommand{\etalchar}[1]{$^{#1}$}
\providecommand{\bysame}{\leavevmode\hbox to3em{\hrulefill}\thinspace}
\providecommand{\MR}{\relax\ifhmode\unskip\space\fi MR }
\providecommand{\MRhref}[2]{%
  \href{http://www.ams.org/mathscinet-getitem?mr=#1}{#2}
}
\providecommand{\href}[2]{#2}
\begin{thebibliography}{CDBDK90}

\bibitem[A{\etalchar{+}}15]{tensorflow2015-whitepaper}
M.~Abadi et~al., \emph{{TensorFlow}: Large-{S}cale {M}achine {L}earning on
  {H}eterogeneous {S}ystems}, 2015, Software available from tensorflow.org.

\bibitem[{\'A}RL12]{Alvarez:2012aa}
M.~A. {\'A}lvarez, L.~Rosasco, and N.~D. Lawrence, \emph{{Kernels for
  Vector-Valued Functions: A Review}}, Foundations and Trends in Machine
  Learning \textbf{4} (2012), no.~3, 195--266.

\bibitem[Bap14]{Bapat:2014aa}
R.~B. Bapat, \emph{{Graphs and Matrices}}, Springer, 2014.

\bibitem[BCR84]{Berg:1984aa}
C.~Berg, J.~P.~R. Christensen, and P.~Ressel, \emph{{Harmonic Analysis on
  Semigroups: Theory of Positive Definite and Related Functions}}, Springer,
  1984.

\bibitem[BDO95]{Berry:1995wp}
M.~W. Berry, S.~T. Dumais, and G.~W. O'Brien, \emph{{Using Linear Algebra for
  Intelligent Information Retrival}}, {SIAM Review} \textbf{37} (1995), no.~4,
  573--595.

\bibitem[BGLV18]{Bachoc:2018aa}
F.~Bachoc, F.~Gamboa, J.-M. Loubes, and N.~Venet, \emph{{A Gaussian Process
  Regression Model for Distribution Inputs}}, IEEE Trans. Information Theory
  \textbf{64} (2018), no.~10, 6620--6637.

\bibitem[BJ08]{Bouhamidi:2008vk}
A.~Bouhamidi and K.~Jbilou, \emph{{A Note on the Numerical Approximate
  Solutions for Generalized Matrix Equations with Applications}}, Appl. Math.
  Comp. \textbf{206} (2008), no.~2, 687--694.

\bibitem[Br{\'e}17]{Bremaud:2017aa}
P.~Br{\'e}maud, \emph{{Discrete Probability Models and Methods}}, Springer,
  2017.

\bibitem[BSdB16]{Brouard:2016aa}
C.~Brouard, M.~Szafranski, and F.~d'Alch{{\'e}} Buc, \emph{{Input Output Kernel
  Regression: Supervised and Semi-Supervised Structured Output Prediction with
  Operator-Valued Kernels}}, Journal of Machine Learning Research \textbf{17}
  (2016), no.~176, 1--48.

\bibitem[BTA04]{Berlinet:2004aa}
A.~Berlinet and C.~Thomas-Agnan, \emph{{Reproducing Kernel Hilbert Spaces in
  Probability and Statistics}}, Springer, 2004.

\bibitem[C{\etalchar{+}}15]{chollet2015keras}
Fran\c{c}ois Chollet et~al., \emph{Keras}, \url{https://keras.io}, 2015.

\bibitem[CDBDK90]{castets1990experimental}
V.~Castets, E.~Dulos, J.~Boissonade, and P.~De~Kepper, \emph{Experimental
  {E}vidence of a {S}ustained {S}tanding {T}uring-{T}ype {N}onequilibrium
  {C}hemical {P}attern}, Physical Review Letters \textbf{64} (1990), no.~24,
  2953.

\bibitem[CFVM19]{campillo2019bayesian}
E.~Campillo-Funollet, C.~Venkataraman, and A.~Madzvamuse, \emph{{Bayesian
  Parameter Identification for Turing Systems on Stationary and Evolving
  Domains}}, Bulletin of Mathematical Biology \textbf{81} (2019), no.~1,
  81--104.

\bibitem[CS01]{cucker2001mathematical}
F.~Cucker and S.~Smale, \emph{On the {M}athematical {F}oundations of
  {L}earning}, Bulletin of AMS \textbf{39} (2001), no.~1, 1--49.

\bibitem[DS84]{Doyle:1984aa}
P.~G. Doyle and J.~L. Snell, \emph{{Random Walks and Electric Networks}},
  Cambridge University Press, 1984.

\bibitem[EMP05]{Evgeniou:2005aa}
T.~Evgeniou, C.~A. Miccelli, and M.~Pontil, \emph{{Learning Multiple Tasks with
  Kernel Methods}}, Journal of Machine Learning Research \textbf{6} (2005),
  615--637.

\bibitem[EOP{\etalchar{+}}12]{economou2012periodic}
A.~D. Economou, A.~Ohazama, T.~Porntaveetus, P.~T. Sharpe, S.~Kondo, M.~A.
  Basson, A.~Gritli-Linde, M.~T. Cobourne, and J.~B.~A. Green, \emph{Periodic
  {S}tripe {F}ormation by a {T}uring {M}echanism {O}perating at {G}rowth
  {Z}ones in the {M}ammalian {P}alate}, Nature Genetics \textbf{44} (2012),
  no.~3, 348--351.

\bibitem[EPP00]{Evgeniou:2000vo}
T.~Evgeniou, M.~Pontil, and T.~Poggio, \emph{{Regularization Networks and
  Support Vector Machines}}, Advances in Computational Mathematics \textbf{13}
  (2000), 1--50.

\bibitem[For10]{Fortunato:2010aa}
S.~Fortunato, \emph{{Community Detection in Graphs}}, Physics Reports
  \textbf{486} (2010), 75--174.

\bibitem[GG96]{gatenby1996reaction}
R.~A. Gatenby and E.~T. Gawlinski, \emph{{A Reaction-Diffusion Model of Cancer
  Invasion}}, Cancer Research \textbf{56} (1996), no.~24, 5745--5753.

\bibitem[GM72]{gierer1972theory}
A.~Gierer and H.~Meinhardt, \emph{{A Theory of Biological Pattern Formation}},
  Kybernetik \textbf{12} (1972), no.~1, 30--39.

\bibitem[GMT10]{garvie2010efficient}
M.~R. Garvie, P.~K. Maini, and C.~Trenchea, \emph{{An Efficient and Robust
  Numerical Algorithm for Estimating Parameters in Turing Systems}}, Journal of
  Computational Physics \textbf{229} (2010), no.~19, 7058--7071.

\bibitem[GT14]{garvie2014identification}
M.~R. Garvie and C.~Trenchea, \emph{{Identification of Space-Time Distributed
  Parameters in the Gierer--Meinhardt Reaction-Diffusion System}}, SIAM Journal
  on Applied Mathematics \textbf{74} (2014), no.~1, 147--166.

\bibitem[GWK{\etalchar{+}}18]{gu2018recent}
J.~Gu, Z.~Wang, J.~Kuen, L.~Ma, A.~Shahroudy, B.~Shuai, T.~Liu, X.~Wang,
  G.~Wang, J.~Cai, et~al., \emph{{Recent Advances in Convolutional Neural
  Networks}}, Pattern Recognition \textbf{77} (2018), 354--377.

\bibitem[HB05]{Hein:2005aa}
M.~Hein and O.~Bousquet, \emph{{Hilbertian Metrics and Positive Definite
  Kernels on Probability Measures}}, Proc. AISTATS, 2005, pp.~136--143.

\bibitem[HJ13]{Horn:2013aa}
R.~A. Horn and C.~R. Johnson, \emph{{Matrix Analysis}}, 2nd ed., Cambridge
  University Press, 2013.

\bibitem[HLBV94]{holmes1994partial}
E.~E. Holmes, M.~A. Lewis, J.~E. Banks, and R.~R. Veit, \emph{{Partial
  Differential Equations in Ecology: Spatial Interactions and Population
  Dynamics}}, Ecology \textbf{75} (1994), no.~1, 17--29.

\bibitem[HR11]{Honeine:2011aa}
P.~Honeine and C.~Richard, \emph{{Preimage Problem in Kernel-Based Machine
  Learning}}, IEEE Signal Processing Magazine \textbf{28} (2011), no.~2,
  77--88.

\bibitem[HSS08]{Hofmann:2008aa}
T.~Hofmann, B.~Sch{\"o}lkopf, and A.~J. Smola, \emph{{Kernel Methods in Machine
  Learning}}, Ann. Statistics \textbf{36} (2008), no.~3, 1171--1220.

\bibitem[JFWJ{\etalchar{+}}98]{jung1998local}
H.-S. Jung, R.~B. Francis-West, P. H. .and~Widelitz, T.-X. Jiang,
  S.~Ting-Berreth, C.~Tickle, L.~Wolpert, and C.-M. Chuong, \emph{{Local
  Inhibitory Action of BMPs and their Relationships with Activators in Feather
  Formation: Implications for Periodic Patterning}}, Developmental Biology
  \textbf{196} (1998), no.~1, 11--23.

\bibitem[KGP13]{Kadri:2013vm}
H.~Kadri, M.~Ghavamzadeh, and P.~Preux, \emph{{A Generalized Kernel Approach to
  Structured Output Learning}}, Proc. Machine Learning Research, vol.~28, 2013,
  pp.~471--479.

\bibitem[KH20]{kazarnikov2020statistical}
A.~Kazarnikov and H.~Haario, \emph{{Statistical Approach for Parameter
  Identification by Turing Patterns}}, Journal of Theoretical Biology
  \textbf{501} (2020), 110319.

\bibitem[KM10]{kondo2010reaction}
S.~Kondo and T.~Miura, \emph{{Reaction-Diffusion Model as a Framework for
  Understanding Biological Pattern Formation}}, Science \textbf{329} (2010),
  no.~5999, 1616--1620.

\bibitem[KR93]{Klein:1993aa}
D.~J. Klein and M.~Randi\'{c}, \emph{{Resistance Distance}}, J. Math. Chemistry
  \textbf{12} (1993), 81--95.

\bibitem[KUK20]{karasozen2020reduced}
B.~Karas{\"o}zen, M.~Uzunca, and T.~K{\"u}{\c{c}}{\"u}kseyhan, \emph{{Reduced
  Order Optimal Control of the Convective FitzHugh--Nagumo Equations}},
  Computers \& Mathematics with Applications \textbf{79} (2020), no.~4,
  982--995.

\bibitem[LJDM20]{landge2020pattern}
A.~N. Landge, B.~M. Jordan, X.~Diego, and P.~M{\"u}ller, \emph{{Pattern
  Formation Mechanisms of Self-Organizing Reaction-Diffusion Systems}},
  Developmental Biology \textbf{460} (2020), no.~1, 2--11.

\bibitem[Mar15]{martcheva2015introduction}
M.~Martcheva, \emph{{An Introduction to Mathematical Epidemiology}}, Text in
  Applied Mathematics, vol.~61, Springer, 2015.

\bibitem[MBM16]{Minh:2016aa}
H.~Q. Minh, L.~Bazzani, and V.~Murino, \emph{{A Unifying Framework in
  Vector-valued Reproducing Kernel Hilbert Spaces for Manifold Regularization
  and Co-Regularized Multi-view Learning}}, Journal of Machine Learning
  Research \textbf{17} (2016), no.~25, 1--72.

\bibitem[MP05]{Micchelli:2005aa}
C.A. Micchelli and M.~Pontil, \emph{{On Learning Vector-Valued Functions}},
  Neural Computation \textbf{17} (2005), 177--204.

\bibitem[Mur82]{murray1982parameter}
J.~D. Murray, \emph{{Parameter Space for Turing Instability in Reaction
  Diffusion Mechanisms: a Comparison of Models}}, Journal of Theoretical
  Biology \textbf{98} (1982), no.~1, 143--163.

\bibitem[Mur01]{murray2001mathematical}
\bysame, \emph{{Mathematical Biology II: Spatial Models and Biomedical
  Applications}}, Springer-Verlag, 2001.

\bibitem[MVM18]{murphy2018parameter}
L.~Murphy, C.~Venkataraman, and A.~Madzvamuse, \emph{{Parameter Identification
  through Mode Isolation for Reaction--Diffusion Systems on Arbitrary
  Geometries}}, International Journal of Biomathematics \textbf{11} (2018),
  no.~04, 1850053.

\bibitem[NTKK09]{nakamasu2009interactions}
A.~Nakamasu, G.~Takahashi, A.~Kanbe, and S.~Kondo, \emph{{Interactions between
  Zebrafish Pigment Cells Responsible for the Generation of Turing Patterns}},
  Proceedings of the National Academy of Sciences \textbf{106} (2009), no.~21,
  8429--8434.

\bibitem[PC19]{peyre2019computational}
G.~Peyr{\'e} and M.~Cuturi, \emph{{Computational Optimal Transport: with
  Applications to Data Science}}, Foundations and Trends{\textregistered} in
  Machine Learning \textbf{11} (2019), no.~5-6, 355--607.

\bibitem[Per15]{Pertham:2015aa}
B.~Pertham, \emph{{Parabolic Equations in Biology}}, Springer, 2015.

\bibitem[PR16]{Paulsen:2016aa}
V.~I. Paulsen and M.~Raghupathi, \emph{{An Introduction to the Theory of
  Reproducing Kernel Hilbert Spaces}}, Cambridge University Press, 2016.

\bibitem[Res21]{Mathematica}
Wolfram Research, \emph{Mathematica, {V}ersion 12.3.1}, 2021.

\bibitem[RMRS14]{raspopovic2014digit}
J.~Raspopovic, L.~Marcon, L.~Russo, and J.~Sharpe, \emph{{Digit Patterning is
  Controlled by a Bmp-Sox9-Wnt Turing Network Modulated by Morphogen
  Gradients}}, Science \textbf{345} (2014), no.~6196, 566--570.

\bibitem[RW09]{Rockafellar:2009aa}
R.~T. Rockafellar and R.~J.-B. Wets, \emph{{Variational Analysis}}, 3rd ed.,
  Springer, 2009.

\bibitem[San15]{Santambrogio:2015aa}
F.~Santambrogio, \emph{{Optimal Transport for Applied Mathematicians}},
  Birkh{\"a}user, 2015.

\bibitem[SC16]{Schaeffer:2016aa}
D.~G. Schaeffer and J.~W. Cain, \emph{{Ordinary Differential Equations: Basics
  and Beyond}}, Springer, 2016.

\bibitem[SHS01]{Scholkopf:2001aa}
B.~Sch\"{o}lkopf, R.~Herbrich, and A.~J. Smola, \emph{{A Generalized
  Representer Theorem}}, Computational Learning Theory, vol. 2111, Springer,
  2001, pp.~416--426.

\bibitem[SLB19]{sgura2019parameter}
I.~Sgura, A.~S. Lawless, and B.~Bozzini, \emph{{Parameter Estimation for a
  Morphochemical Reaction-Diffusion Model of Electrochemical Pattern
  Formation}}, Inverse Problems in Science and Engineering \textbf{27} (2019),
  no.~5, 618--647.

\bibitem[SMB{\etalchar{+}}99]{Scholkopf:1999aa}
B.~Sch\"{o}lkopf, S.~Mika, C.~J.~C. Burges, P.~Knirsch, K.-R. M\"{u}ller,
  G.~R\"{a}tsch, and A.~J. Smola, \emph{{Input Space Versus Feature Space in
  Kernel-Based Methods}}, IEEE Trans. Neural Networks \textbf{10} (1999),
  no.~5, 1000--1017.

\bibitem[SPM16]{stoll2016fast}
M.~Stoll, J.~W. Pearson, and P.~K. Maini, \emph{{Fast Solvers for Optimal
  Control Problems from Pattern Formation}}, Journal of Computational Physics
  \textbf{304} (2016), 27--45.

\bibitem[SRTS06]{sick2006wnt}
S.~Sick, S.~Reinker, J.~Timmer, and T.~Schlake, \emph{{WNT and DKK Determine
  Hair Follicle Spacing through a Reaction-Diffusion Mechanism}}, Science
  \textbf{314} (2006), no.~5804, 1447--1450.

\bibitem[SS04]{Smola:2004aa}
A.~J. Smola and B.~Sch\"{o}lkopf, \emph{{A Tutorial on Support Vector
  Regression}}, Statistics and Computing \textbf{14} (2004), 199--222.

\bibitem[SS20]{shangerganesh2020optimal}
L.~Shangerganesh and P.~T. Sowndarrajan, \emph{{An Optimal Control Problem of
  Nonlocal Pyragas Feedback Controllers for Convective FitzHugh--Nagumo
  Equations with Time-Delay}}, SIAM Journal on Control and Optimization
  \textbf{58} (2020), no.~6, 3613--3631.

\bibitem[SSIS19]{scholes2019comprehensive}
N.~S. Scholes, D.~Schnoerr, M.~Isalan, and M.~P.~H. Stumpf, \emph{{A
  comprehensive network atlas reveals that Turing patterns are common but not
  robust}}, Cell Systems \textbf{9} (2019), no.~3, 243--257.

\bibitem[SST14]{Seto:2014aa}
M.~Seto, S.~Suda, and T.~Taniguchi, \emph{{Gram Matrices of Reproducing Kernel
  Hilbert Spaces over Graphs}}, Linear Algebra and its Applications
  \textbf{445} (2014), 56--68.

\bibitem[TCP{\etalchar{+}}18]{tan2018polyamide}
Z.~Tan, S.~Chen, X.~Peng, L.~Zhang, and C.~Gao, \emph{{Polyamide Membranes with
  Nanoscale Turing Structures for Water Ourification}}, Science \textbf{360}
  (2018), no.~6388, 518--521.

\bibitem[Tur52]{turing1952chemical}
A.~Turing, \emph{{The Chemical Basis of Morphogenesis}}, Philosophical
  Transactions of the Royal Society of London B \textbf{237} (1952), no.~641,
  37--72.

\bibitem[UKYK17]{uzunca2017optimal}
M.~Uzunca, T.~K{\"u}{\c{c}}{\"u}kseyhan, H.~Y{\"u}cel, and B.~Karas{\"o}zen,
  \emph{{Optimal Control of Convective FitzHugh--Nagumo Equation}}, Computers
  \& Mathematics with Applications \textbf{73} (2017), no.~9, 2151--2169.

\bibitem[Wha90]{Whaba:1990uc}
G.~Whaba, \emph{{Spline Models for Observational Data}}, SIAM, 1990.

\bibitem[WKG21]{woolley2021bespoke}
T.~E. Woolley, A.~L. Krause, and E.~A. Gaffney, \emph{{Bespoke Turing
  Systems}}, Bulletin of Mathematical Biology \textbf{83} (2021), no.~5, 1--32.

\end{thebibliography}

\end{document}